\documentclass{article}

\usepackage[final]{neurips_2020}

\usepackage[utf8]{inputenc} 
\usepackage[T1]{fontenc}    
\usepackage{hyperref}       
\usepackage{url}            
\usepackage{booktabs}       
\usepackage{amsfonts}       
\usepackage{nicefrac}       
\usepackage{microtype}      
\usepackage{graphicx}       
\usepackage{float}
\usepackage{amsmath}
\usepackage{fancyvrb}
\usepackage{fvextra}
\usepackage{array}

\usepackage{tikz, pgfplots}
\usetikzlibrary{positioning}
\usetikzlibrary{arrows.meta}

\usepackage{caption}

\usepackage{xcolor}         
\hypersetup{
    colorlinks,
    linkcolor={black!50!black},
    citecolor={blue!50!black},
    urlcolor={blue!80!black}
}
\setcitestyle{authoryear,open={(},close={)}}

\title{Towards Fine-Dining Recipe Generation with Generative Pre-trained Transformers}

\author{%
  Konstantinos Katserelis\\
  Department of Informatics\\
  Athens University of Economics and Business\\
  Athens, Greece \\
  \texttt{konnoskkats@gmail.com} \\
  \small{(p3170065@aueb.gr)} \\
}

\pgfplotsset{compat=1.16}

\begin{document}

\maketitle

\begin{center}
{\bf Supervisor:}\\
Konstantinos Skianis, Ph.D. \\
Department of Informatics\\
  Athens University of Economics and Business\\
  Athens, Greece \\
  \texttt{skianis.konstantinos@gmail.com} \\
\vspace{1cm}
    \resizebox{\textwidth}{!}{\includegraphics{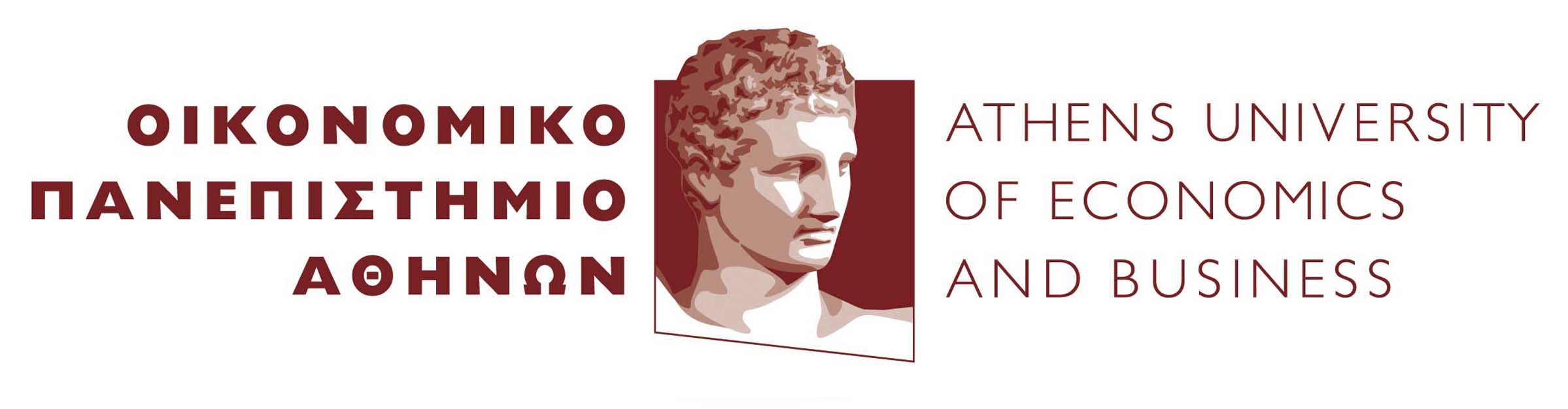}}
    \vspace{1cm}
\end{center}

\begin{abstract}
Food is essential to human survival. So much so that we have developed different recipes to suit our taste needs. In this work, we propose a novel way of creating new, fine-dining recipes from scratch using Transformers, specifically auto-regressive language models. Given a small dataset of food recipes, we try to train models to identify cooking techniques, propose novel recipes, and test the power of fine-tuning with minimal data.
Code and data can be found   \href{https://github.com/its-kos/Fine-Dining-Recipe-generation}{here}.
\end{abstract}

\clearpage

\tableofcontents

\clearpage

\listoffigures
\listoftables

\clearpage

\section*{Acknowledgements}

I would like to acknowledge and express my special thanks to my supervisor Konstantinos Skianis for the opportunity to work on this project and for his time in helping and guiding me through this brand new experience, as well as my parents and friends for enduring me during this time.

\clearpage

\section{Introduction}
Automatic cooking recipe generation is an intriguing and useful research question that can assist with overcoming the drawbacks of conventional recipe retrieval systems.

All recipes, no matter their contents and level of difficulty, follow a certain structure to an extend. It includes the recipe name, the ingredients and the instructions. All the recipes we will use will include this information. On top of that, recipes can be seen as a sequence of characters which makes them great inputs to character-level Recurrent neural networks as well as autoregressive models like the GPT-2 transformer. We would like to train the models on existing recipes and then have them recommend brand new ones. \par

In the following experiments, we will make use of a transformer, more specifically a pre-trained \href{https://en.wikipedia.org/wiki/GPT-2}{\emph{GPT-2}} model in order to generate structured recipes from scratch. For all this, we are going to be using a brand new dataset we created, from publicly available data. The novel dataset we create features specifically "fine-dining" recipes and the model is specifically fine-tuned to generate "gourmet" dishes unlike all the previous "normal" recipe models.


\section{Background}

\subsection{Machine Learning}
Machine learning (ML) is a topic of study focused on comprehending and developing "learning" methods, or methods that use data to enhance performance on a certain set of tasks \citep{mitchell1997machine}.
It is considered to be a component of artificial intelligence.
In order to generate predictions or choices without being explicitly taught to do so, machine learning algorithms construct a model from sample data, often known as training data.
Machine learning algorithms are utilized in a wide range of applications, including speech recognition, email filtering, computer vision, and medicine, when it is challenging or impractical to create traditional algorithms to carry out the required functions.

\subsection{Neural Networks and Deep Learning}

Artificial neural networks (ANNs), more commonly known as neural networks (NNs) or even just "neural nets", are computer architectures comprised of numbers and activation functions that take their design cues from the biological neural networks that make up brains.

Artificial neurons, which are a set of interconnected units or nodes that loosely resemble the neurons in a biological brain, are the foundation of an ANN.
Like the synapses in a human brain, each link has the ability to send a signal to neighboring neurons.
An artificial neuron can signal neurons that are connected to it after processing signals that are sent to it.
Each neuron's output is calculated by some non-linear function of the sum of its inputs, and the "signal" at each link is a real integer.
Edges are what link the points.
Typically, neurons and edges have an adjustable weight.

Deep learning, commonly referred to as deep structured learning, is one of several machine learning techniques built on representation learning and artificial neural networks \citep{lecun2015deep, goodfellow2016deep}.
As in machine learning, it includes unsupervised, semi-supervised, and supervised learning methods.

In fields like computer vision \citep{krizhevsky2017imagenet}, speech recognition \citep{deng2013new}, natural language processing \citep{mikolov2013efficient, goldberg2016primer}, machine translation \citep{bahdanau2014neural}, bioinformatics \citep{min2017deep}, drug design \citep{jing2018deep}, medical image analysis \citep{litjens2017survey}, climate science \citep{rasp2018deep}, protein structure prediction \cite{jumper2021highly}, and board game programs \citep{silver2017mastering}, deep-learning architectures like deep neural networks, deep belief networks, deep reinforcement learning, recurrent neural networks, and convolutional neural networks have been used.
These applications have led to results that are comparable to and in some cases even better than those of traditional approaches.

The information processing and distributed communication nodes in biological systems served as the inspiration for artificial neural networks (ANNs).

\subsubsection{Recurrent Neural Networks (RNNs)}
Recurrent neural networks are a general type of neural network which allow storing previous outputs as a "\textit{hidden state}" and using that as input in the next iteration. This seemingly simple strategy allows them to form a form of temporal dynamic behavior and makes them suitable for tasks which include sequences with context such as speech, text etc. \par
The strength of \textit{character-level} Recurrent Neural Networks has long been outlined by \cite{karpathy2015unreasonable}.
So that's what we started looking at at first.

\subsubsection{Generative Adversarial Networks (GANs)}
Generative Adversarial Networks (more commonly known as "\textit{GANs}") are a type of neural network architecture firstly designed and proposed by \cite{https://doi.org/10.48550/arxiv.1406.2661}.
In this type of architecture, two neural networks (A \textit{Generator} and a \textit{Discriminator}) compete in a zero-sum game of "cat and mouse". Given a training dataset, the goal of the \emph{Generative} model is to try and learn how to generate examples with the same statistics as the given data. Then, these examples are passed to the \emph{Discriminator} whose job is to predict whether the data came from the dataset or the \emph{Generator}.
Once the training finishes, the \emph{Generator} should be able to produce data that "fool" the \emph{Discriminator} and thus might be able to perform tasks such as generating "realistic" paintings, photos etc. \par
In our case, a possible use for GANs is generating recipe images so the user knows what the dish might look like in the end. For this reason, later on we explore datasets which include images as well as text.

\subsubsection{Convolutional Neural Networks (CNNs)}
Convolutional neural networks, (also known as CNNs) are a type of neural net commonly used for image analysis and identification \citep{lecun1995learning}.
The name stems from the use of filters which "convolve" over the image (Even though there is an on-going debate over the name of this action. Some people claim it's actually cross-corellating rather than convolve). The "deeper" filters, gradually learn features of the image (such as edges, corners, lines, etc) which are then combined on "shallower" filters to form more complex features which are then fed into traditional neural networks. This makes CNNs ideal for for image analysis and no so much for generation such as GANs. \par
In our project, they can be used to identify recipe ingredients, styles and cooking methods.

\subsubsection{Transformers}
Much like Recurrent Neural Networks and LSTM (Long Short Term Memory) architectures, \textit{Transformers} are a deep learning model which deploy the strategy of "\href{https://en.wikipedia.org/wiki/Attention_(machine_learning)}{self-attention}" during the learning process.
In the paper by \cite{https://doi.org/10.48550/arxiv.1706.03762}, \textit{Transformers} are introduced as an alternative to the recurrent and convolution methods found in RNNs and CNNs.

\paragraph{GPT2}
OpenAI developed an open-source artificial intelligence known as Generative Pre-trained Transformer 2 (GPT-2) in February 2019.
While sometimes indistinguishable from that of humans, GPT-2 translates text, responds to inquiries, summarizes passages \citep{hegde2020unsupervised}, and provides text output on a level, which can become monotonous or incomprehensible when generating extended passages.
It is a general-purpose learner; none of these activities were particularly taught to it, and its capacity to carry them out is an extension of its general capacity to precisely synthesize the subsequent item in any given sequence \citep{radford2019language}.
As a "direct scale-up" of OpenAI's 2018 GPT model, GPT-2 was developed with a ten-fold increase in both the number of parameters and the size of the training dataset.

The GPT design replaces earlier recurrence- and convolution-based systems with a deep neural network, specifically a transformer model.
The model may selectively focus on input text sections that it thinks to be the most pertinent thanks to attention mechanisms.
This model beats earlier benchmarks for RNN/CNN/LSTM-based models and enables far more parallelization.

In November 2019, OpenAI made available the whole GPT-2 language model (with 1.5 billion parameters).

The 175 billion parameter GPT-3, which was scheduled to be released to the public in 2020, was to follow GPT-2 (whose source code has never been made available).
GPT-3 can only be accessed via an API that Microsoft provides.

\paragraph{T5}
In 2020, in the paper by \cite{t5google}, Google built upon the concept of "transfer-learning".
Essentially using a model that is pre-trained on a "data-rich task" and fine-tuning it to a different one.
Through this powerful technique, Google proposed a unified framework for NLP projects which converts all text problems into a text-to-text problem and published the both the models and data, along them was the T5 transformer.
A pre-trained encoder-decoder which transforms the problems at hand into text-to-text ones and works really well without the need for any further adjustment on a plethora of problems.
T5 transformers come at the following sizes:

\begin{itemize}
    \item t5-small
    \item t5-base
    \item t5-large
    \item t5-3b
    \item t5-11b.
\end{itemize}

\subsection{Fine-Tuning}
Fine-tuning is a part of transfer learning that is used to specialize pre-trained models, trained on large amounts of data and general problems, to more specific ones. A lot of the times, a smaller amount of data can be used especially when the problem at hand is similar to the ones used in the pre-training phase. This is a benefit of transfer learning, since the models can keep most of the "knowledge" they built at first, and adapt it to the new data. An example of transfer-learning is that of specializing CNNs for object detection. Since "deeper" layers learn how to identify just shapes and edges which are common on all objects, we can remove "higher" layers and re-train the model to identify more specialized objects without the need for the model to re-learn how to identify shapes, edges, etc. \par
In our case, we will be mostly using fine-tuning to train models because the task of recipe generation can been seen simply as that of text generation. In essence, given (a) string(s) (keywords, ingredients), we want to be able to generate a sequence of text which hopefully will be a recipe. On top of that, as mentioned in the "Data" section our dataset was really limited due to the lack (or rather absence) of "fine-dining" datasets and even websites. Thus, a pre-trained model will have already been trained on how to generate "meaningful" sequences. We just have to specialize it into making ones that turn out to be recipes. To help us with this task, we turn to \href{https://huggingface.co/}{Hugging Face}. It is a popular community and data science platform that provides users with tools, models and data as well as a share repository or models and knowledge for everyone to access (a lot of times at a price).

\section{Related work}

Several methods for producing recipes text have been put forth, including knowledge-based models \citep{varshney2019big} and deep neural network models \citep{majumder-etal-2019-generating, salvador2019inverse}.

\subsection{RecipeGAN}
RecipeGAN was an attempt to create a Generative Adversarial Network \cite{https://doi.org/10.48550/arxiv.1406.2661} and more specifically a Tabular Generative Adversarial Neural Network (TGAN) to create new recipes trained on recipes sourced from the internet. It also contains functionality to compute the nutrition data for each recipe using USDA official nutrition data and Natural Language Processing (NLP), and fuzzy logic for each ingredient in the recipe. The data for this project came from 2 sources, namely the websites \href{https://www.allrecipes.com/}{AllRecipes} which features a large collection of everyday recipes whose data and images were used and \href{https://fdc.nal.usda.gov/}{FoodData Central} which contains every food registered with the USDA (U.S department of agriculture) and their nutrition contents. \par
However, the idea used in this project is different from ours since \emph{RecipeGAN} aims to create images from ingredients rather than the other way around and also it includes calculating nutritional data for each recipe.

\subsection{RecipeNLG}
RecipeNLG was the product that came out by \cite{bien-etal-2020-recipenlg}, in which a T5 transformer (named "Chef Transformer") was trained on 2,231,142 recipes which can be found in the \href{https://recipenlg.cs.put.poznan.pl/}{RecipeNLG} dataset. In this paper, the same problems we faced are outlined. Specifically, it is mentioned how the absence of suitable data hinders the use of state of the art model as well as the fact that many datasets are created with computer vision goals in mind. Both of these problems were the main difficulties we faced.

Like we mentioned, they set certain "rules" which all recipes should follow, like the structure and the contents of them. The main bulk of the work was creating and processing the dataset into a proper, usable format. Furthermore, on top of the pre-trained \emph{GPT-2}, a \emph{NER} (Named Entity Recognizer) was used, taught the ingredients and used to discern between food and ingredient entities.

\subsection{Transformers}
Recently, it has been demonstrated that large-scale transformer-based language models outperform recurrent neural networks (RNNs) in a number of natural language processing (NLP) tasks.
Transformers are renowned in text creation for their efficiency in capturing complicated relationships and producing concise phrases.
Among these, OpenAI's GPT-2 has demonstrated outstanding performance in a range of text production tasks \citep{radford2019language} after being pre-trained on a gigaword-scale textual dataset.
A recent study has also demonstrated that fine-tuning GPT-2 can improve performance on text creation for specific domains \citep{zhang2019dialogpt}.
However, pre-trained transformer-based language models' efficacy in generating cooking recipes has not yet been investigated.

\section{Data}
\label{sec:data}
As mentioned before, our data should consist of (preferably fine-dining) recipes which is what we are trying to generate. Minimum, they must include the \emph{title}, \emph{ingredients} and \emph{instructions}. It is also possible to include dish images for dish image generation. Possible data sources that we explored were the following:

\subsection{General Recipe Datasets}

\begin{itemize}
    \item \href{http://pic2recipe.csail.mit.edu/}{\textbf{Recipe1M}}
    \item \href{https://eightportions.com/datasets/Recipes/}{\textbf{Recipe Box}}
    \item \href{https://recipenlg.cs.put.poznan.pl/}{\textbf{RecipeNLG}} | Hugging Face
    \item \href{https://www.kaggle.com/datasets/kaggle/recipe-ingredients-dataset}{\textbf{Recipe Ingredients}} | Kaggle
    \item \href{https://www.kaggle.com/datasets/shuyangli94/food-com-recipes-and-user-interactions?select=RAW_recipes.csv}{\textbf{Food.com Recipes}} | Kaggle
    \item \href{https://www.kaggle.com/datasets/jtrofe/beer-recipes}{\textbf{Beer Recipes}} | Kaggle
    \item \href{https://www.kaggle.com/datasets/hugodarwood/epirecipes?select=full_format_recipes.json}{\textbf{Epicurious}} | Kaggle
\end{itemize}

\subsubsection{Recipe1M}

Recipe1M is a dataset introduced in \cite{marin2019recipe1m+}, consisting of over one million cooking recipes and 13 million food images. It was firstly used to learn cross-model embeddings, and its main purpose was to train a neural network on an image-recipe retrieval task. \\
The dataset consists of the following files:

\begin{description}
    \item[Training set] - 95GiB
    \item[Validation set] - 21 Gib
    \item[Test set] - 20 GiB
\end{description}

On top of that, there are 16 (210 GiB) tar files with recipe images. The creators also include model training files such as

\begin{description}
    \item[data.h5.gz] - Model parameters
    \item[vocabv.bin.gz] - Vocabulary of the dataset
    \item[encs\_train\_1024.t7] - Train set encodings
    \item[encs\_val\_1024.t7] - Validation set encodings
    \item[encs\_test\_1024.t7] - Test set encodings
\end{description}

\subsubsection{RecipeNLG}
\href{https://huggingface.co/}{Hugging Face} created this dataset \citep{bien-etal-2020-recipenlg} of the Flax/Jax Community Week and it includes 2,231,142 recipes. It was published along an interactive app for recipe/text generation.

\subsubsection{Recipe Ingredients}

This dataset comes in the form of 2 files. A "train" and a "test" file which include recipe ids, type of cuisine, and list of ingredients. The data was taken from \href{https://www.yummly.com/}{Yummly}.

\subsubsection{Food.com}

Food.com contains 180000+ recipes and 700000+ reviews coming from \href{https://www.food.com/}{Food.com}.
The data was used in the paper by \cite{majumder-etal-2019-generating}.

\subsubsection{Beer Recipes}

This is a user generated dataset, downloaded from the website \href{https://www.brewersfriend.com/search/}{Brewers Friend} where user can post their own reviews and beer recipes. It includes 75000 homebrewed beers with 176+ styles. It also includes user reports and details about the beers.

\subsubsection{Epicurious}

The data came from the \href{https://www.epicurious.com/recipes-menus}{Epicurious} website and it includes over 20000 recipes and information like rating, nutritional information and category.

\subsubsection{Recipe Box}

The Recipe Box dataset (and the one we've used for this project) contains roughly 125000 recipes from food websites like \href{https://www.foodnetwork.com}{Food Network}, \href{http://allrecipes.com}{AllRecipes} and \href{http://www.epicurious.com}{Epicurious}. The recipes include a title, ingredients and measurements, instructions and some of the recipes, a picture of the resulting dish. The files provided are

\begin{description}
    \item[recipes\_raw\_nosource\_ar.json] - 47,4 MB, 39802 recipes - Recipes from Allrecipes website
    \item[recipes\_raw\_nosource\_epi.json] - 58,3 MB, 25323 recipes - Recipes from Epicurious website
    \item[recipes\_raw\_nosource\_fn.json] - 89,3 MB, 60039 recipes - Recipes from Foodnetwork website
\end{description}

The recipe images are not provided as a ready dataset but a script to scrape them is provided \href{https://github.com/rtlee9/recipe-box}{here}.

When merged into a single, \textit{125164} recipe dataset. The structure of each entry (recipe) is the following:

\begin{verbatim}
    <RECIPE-ID>: {
        "title": <RECIPE-TITLE>,
        "ingredients": [
          "<INGREDIENT-AMOUNT> <INGREDIENT-#1>",
          "<INGREDIENT-AMOUNT> <INGREDIENT-#2>",
                        ...
          "<INGREDIENT-AMOUNT> <INGREDIENT-#N>"
        ],
        "instructions": <INSTRUCTIONS-AS-PLAINTEXT>,
        "picture_link": <PICTURE-ID>
    }
\end{verbatim}

We noticed that there were some "anomalies" within the dataset. In various places within the recipes there was the word "ADVERTISEMENT". Also some recipes didn't have any of the "Title", "Ingredients" or "Instructions" labels.

\begin{description}
    \item[Starting recipes] - 125164
    \item[Recipes after validating] - 122938
    \item[Number of incomplete recipes] - 2226
\end{description}

\begin{figure}[H]
\centering
\includegraphics[scale=0.8]{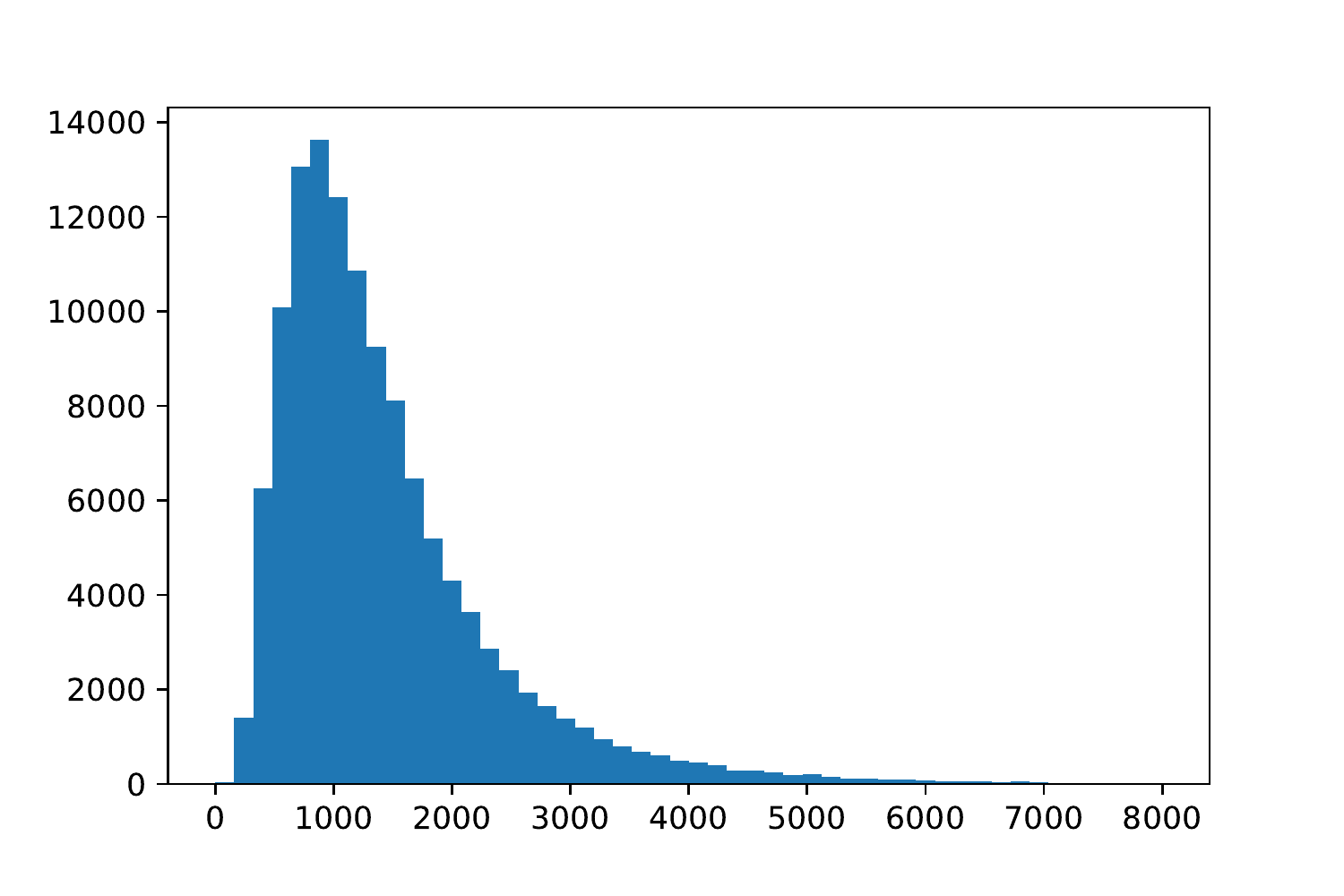}
\caption{Recipe lengths (In total characters)}
\label{rl}
\end{figure}

\begin{figure}[H]
\centering
\includegraphics[scale=0.8]{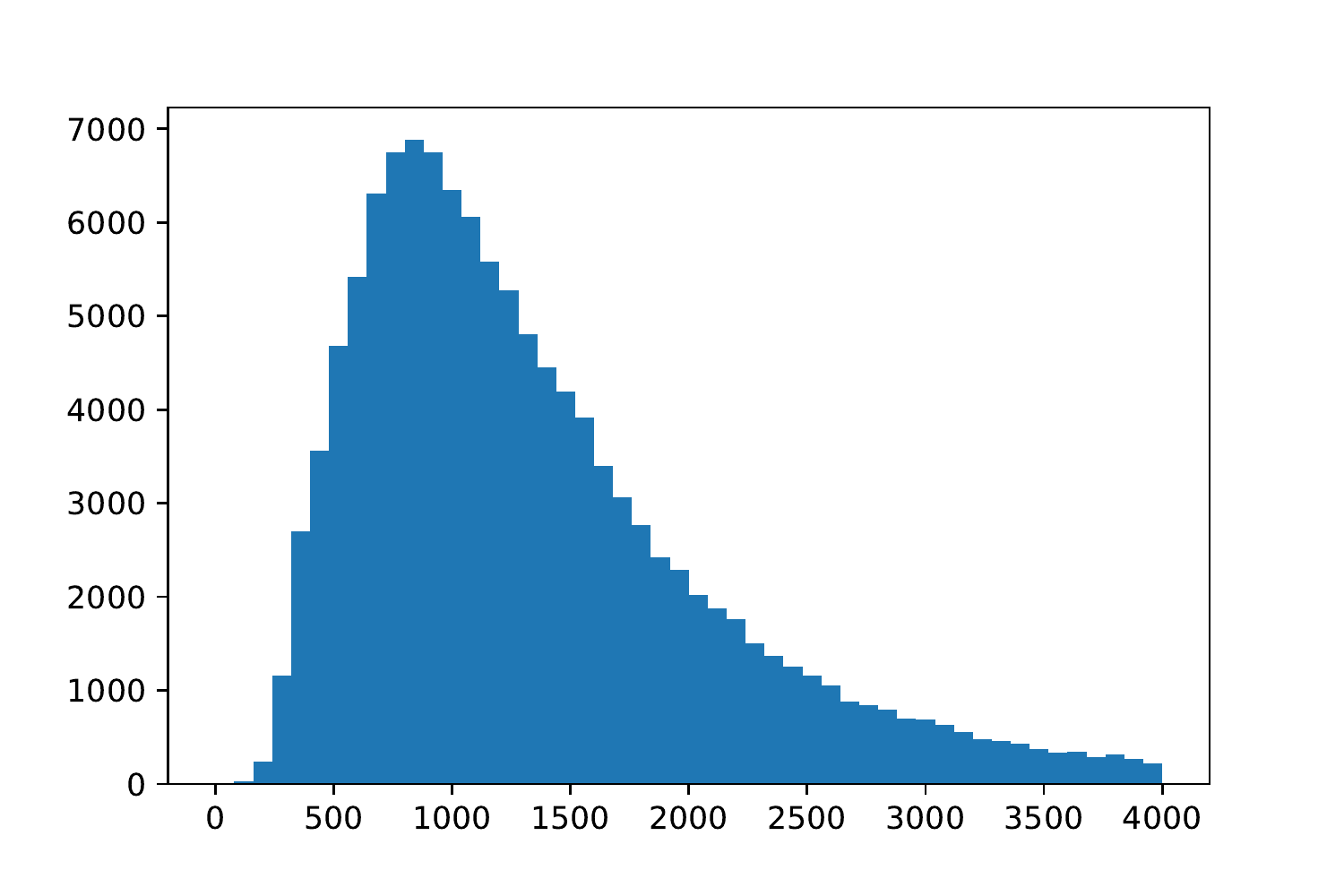}
\caption{Recipe lengths (More precise range)}
\label{rl-mp}
\end{figure}

Furthermore, some of the recipes are over 6000 characters. In the figure \ref{rl} below the distribution of lengths can be seen. In the figure \ref{rl-mp} a more precise scale is used.

By having a closer look its clear that over 75\% of the recipes are under 2000 characters, as shown in Table \ref{stats}.

\begin{table}[h!]
\centering
\begin{tabular}{||c c c c c||}
 \hline
 Min & 25th percentile & 50th percentile & 75th percentile & Max \\ [0.5ex]
 \hline\hline
 74 & 798 & 1185 & 1777.75 & 29988\\ \hline
\end{tabular}
\vspace{0.2cm}
\caption{Recipe Lengths percentiles}
\label{stats}
\end{table}

Thus, when we remove all recipes over 2000 characters long we are left with 23880 usable recipes for learning.

\begin{description}
    \item[Starting recipes] - 122938
    \item[Recipes after validating] - 99058
    \item[Number of removed recipes] - 23880
\end{description}

\subsection{Fine Dining Dataset}
Since our project is aimed towards fine dining rather than general recipe generation we had to have a more thorough look at datasets offering fine dining recipes, however none were aimed towards that so we had to create our own. Some websites that offer fine dining recipes that we used were:

\begin{itemize}
    \item \href{https://www.greatbritishchefs.com/}{\textbf{Great British Chefs}}
    \item \href{https://www.finedininglovers.com}{\textbf{Fine Dining Lovers}}
\end{itemize}

None of these websites offer a public API or any of its data so our only way of acquisition was web-scraping to gather the data and formatting them ourselves into a usable dataset.

\subsubsection{Web Scraping}
\begin{figure}[H]
\centering
\includegraphics[width=\textwidth]{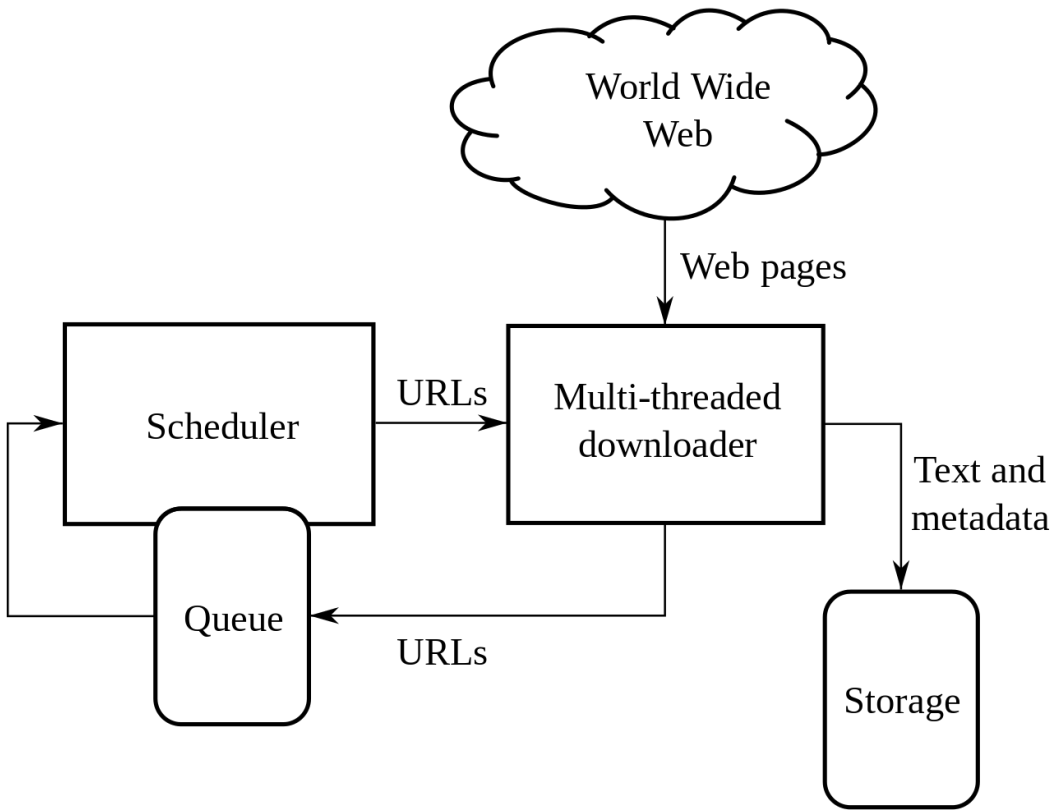}
\caption{General web crawler architecture \citep{webcrawling}.}
\label{webc-arhitecture}
\end{figure}

Web scraping, is the action of "scraping" data from the web through the use of (almost always) automated software. This software can either directly access web pages through the \href{https://en.wikipedia.org/wiki/Hypertext_Transfer_Protocol}{\emph{HTTP}} protocol, or use a standard web browser. The software that does this scraping is often called "\textit{bots}" or more commonly "\textit{web crawlers}". The typical process involves "fetching" (i.e. downloading) a web page, then reading through the contents (most of the time \href{https://en.wikipedia.org/wiki/HTML}{\emph{HTML}} or \href{https://en.wikipedia.org/wiki/XML}{\emph{XML}}) and extracting the information it wants to keep. The figure \ref{webc-arhitecture} displays a general web-crawler architecture. Automated web scraping often involves requesting a web page multiple times which can lead to having the IP of the bot blocked, the API (if any used) disabled and even "drowning" the page, causing an unwanted \href{https://en.wikipedia.org/wiki/Denial-of-service_attack}{DOS} attack.

Another important topic that needs to be covered is that of \emph{sitemaps}. Almost every single web page on the internet has one. Sitemaps are (almost always) XML documents which indicate the location of all the pages, files and valuable information of your web page. They are often found in the following file:

\begin{verbatim}
    www.webpage.com/sitemap.xml
\end{verbatim}

path. Alternatively, many webpages include the location of their sitemap in the \textit{robots.txt} file (meant for crawlers and bots) here:

\begin{verbatim}
    www.webpage.com/robots.txt
\end{verbatim}

For our project, we decided to code a web crawler of our own in \emph{Python}. The 2 most useful python libraries for web crawling are:

\begin{enumerate}
    \item \textbf{Beautiful Soup}
    \item \textbf{Selenium}
\end{enumerate}

\paragraph{Beautiful Soup}
Beautiful soup (aka BS4) is probably the simplest and most straightforward way of pulling data from \textit{HTML} and \textit{XML}. The main 3 advantages of BS4 which can also be found in the \href{https://www.crummy.com/software/BeautifulSoup/}{documentation} are:

\begin{enumerate}
    \item It provides a few simple methods and Pythonic idioms for navigating, searching, and modifying a parse tree: a toolkit for dissecting a document and extracting what you need. It doesn't take much code to write an application
    \item It automatically converts incoming documents to Unicode and outgoing documents to UTF-8. You don't have to think about encodings, unless the document doesn't specify an encoding and Beautiful Soup can't detect one. Then you just have to specify the original encoding.
    \item It sits on top of popular Python parsers like lxml and html5lib, allowing you to try out different parsing strategies or trade speed for flexibility.
\end{enumerate}

Beautiful soup works by converting the document into a "soup" which is a complex tree of Python objects. Then it searches through that to find the required tags and elements. Even though Beautiful soup is really fast, we are looking for something more flexible and something which includes \href{https://en.wikipedia.org/wiki/XPath}{XPath} support like Selenium which we describe below.

\paragraph{Selenium}
Selenium's primary purpose is automating tasks which involve web browsers and web pages, mostly for testing purposes. However, this functionality can easily be used for our purposes and so we move on with Selenium. However, Selenium requires a few extra components to function properly as it works more like a framework rather as a ready-to-go library. It first of all requires the presence of a browser (even if you decide to use headless Selenium), the \emph{Webdriver}, and the \emph{Driver} which is often created and distributed by browser vendors. Selenium uses the webdriver to interface and communicate with the driver which in turn communicates with the browser as can be seen in the image \ref{comms} below:

\begin{figure}[H]
    \centering
    \includegraphics[scale=0.5]{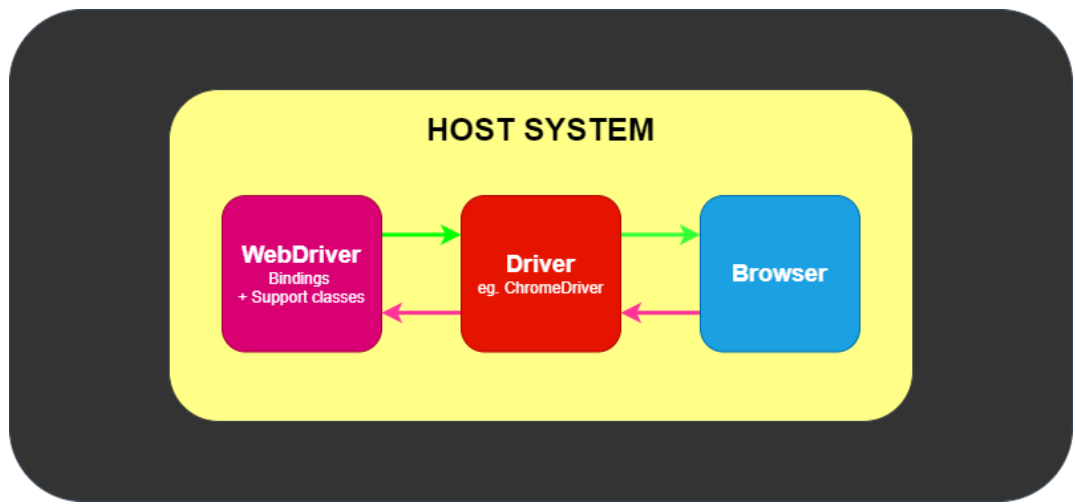}
    \caption{Basic Selenium communications architecture}
    \label{comms}
\end{figure}

Using this, it can communicate with any browser and allows for seamless user simulation. Thus, we use selenium alongside "Google Chrome" for our scraping needs. Furthermore, we decided to use "Google Colab" since one of the features it includes is restarting and changing VMs on the spot. That enables us to get a "fresh" IP every time which is usefull since multiple requests on the same page can result in an IP ban. Below are the options (Selenium accepts certain options that dictate how it will operate) we passed to the driver:

\begin{itemize}
    \item \textbf{--headless} | Headless use dictates to the driver that we do not require any graphical interface of the browser and everything happens in the background. This results in faster operations. This is a required option by Colab.
    \item \textbf{--no-sandbox} | This tells Selenium to not use a sandboxed (C++ library) environment since it can lead to bugs.
    \item \textbf{--disable-dev-shm-usage} | Since our script runs on a Colab Linux VM, using the dev \textbackslash shm folder can lead to bugs so we disable it. This is a required option by Colab
\end{itemize}

\subsection{Dataset creation}
The first step to creating our dataset was gathering all the useful links, i.e. every link from the websites mentioned above that has a recipe. During this step we noticed that for both the websites, the urls of recipes included the folowing String \emph{".com/recipe/"}. A clever way to use that is pair every url found in the sitemaps and check if it includes this String, if yes it's a recipe url -keep it, otherwise it's not -throw it away. The sitemaps are \textit{JSON} files with the following format:

\begin{verbatim}
    <urlset xmlns="http://www.sitemaps.org/schemas/sitemap/0.9">
    <url>
        <loc>https://www.finedininglovers.com/login</loc>
        <priority>0.5</priority>
    </url>
    <url>
        <loc>https://www.finedininglovers.com/downloadable/...</loc>
        <lastmod>2021-09-01T18:48:45+02:00</lastmod>
        <changefreq>weekly</changefreq>
        <priority>0.5</priority>
    </url>
                ...
    </urlset>
\end{verbatim}

Essentially a nested list of all of the website's urls and their attributes. These attributes can include:

\begin{itemize}
    \item \textbf{<loc>} - URL of the page.
    \item \textbf{<lastmod>} - The date of last modification of the page.
    \item \textbf{<changefreq>} - How frequently the page is likely to change.
    \item \textbf{<priority>} - The priority of this URL relative to other URLs on your site. Valid values range from 0.0 to 1.0.
\end{itemize}

(This information come directly from the \href{https://www.sitemaps.org/protocol.html}{page for sitemaps protocols}) \\
(We are only interested in the \emph{<loc>} attribute)

The webpage \href{https://www.finedininglovers.com}{\textbf{Fine Dining Lovers}} has a total of 3838 urls out of which 2272 are recipe ones ($\sim$59\%) \\
\href{https://www.greatbritishchefs.com/}{\textbf{Great British Chefs}} has 9896 total urls out of which 5628 are valid recipe ones ($\sim$56.8\%) \\
Total recipe links found: 7900.

In this part of our project we came along the issue of connection timeouts, really log webpage fetch times and connection refusals. A lot of pages took way too long (some pages took $\sim$100-120 sec. to properly load) and so online scraping was not a viable option. Thus, using the \href{https://vovsoft.com/software/batch-url-downloader/#:~:text=Vovsoft\%20Batch\%20URL\%20Downloader\%20is,part\%20of\%20the\%20same\%20job.}{VOVSOFT batch URL downloader} we downloaded as ".html" files as many of the pages as we could for offline scraping. The same \emph{Read timed out} and \emph{Connect timed out} appeared on some of the pages so we managed to get the following pages:

\begin{center}
\begin{table}
\resizebox{\textwidth}{!}{
\begin{tabular}{|c|c|c|c|c|}
 \hline
 Website & Total URLs & Successfully downloaded & Success & Time \\
 \hline\hline
 \href{https://www.finedininglovers.com}{Fine Dining Lovers} & 2272 & 2268 & 99.8\% & $\sim$11 mins\\
 \href{https://www.greatbritishchefs.com/}{Great British Chefs} & 5628 & 4671 & 83\% & $\sim$140 mins\\
 \hline
\end{tabular}}
\end{table}
\end{center}

This tool downloads the page as a generic file without extension and so we used the following bash script to make them into \textit{html} files

\begin{verbatim}
    for i in $( ls ); do mv $i $i.html; done
\end{verbatim}

For each of these 7900 (possible) downloaded recipe pages, we need to filter out pages that are just recipe styles/cuisines/categories and not actual recipes and ones that are broken/require login to view. In this task we will make use of \textit{XPaths}. It stands for "XML Path language" and it is a way to "traverse" the DOM (i.e. the node tree that is created by the page's HTML) and locate specific elements / nodes of the page. \textit{XPaths} have the following form:

\begin{verbatim}
    Xpath=//tagname[@attribute='value']
\end{verbatim}

Using these, we can search for and locate specific elements on a page which will help us identify pages that are irrelevant and/or broken. To be more specific, we are checking the category link that's located on the page and it is in the following list, we can safely discard the page.

\begin{itemize}
    \item \textbf{Recipe Category} - Describes an entire category of recipes.
    \item \textbf{Cuisine} - Describes an entire cuisines.
    \item \textbf{Cooking Method} - Describes an general cooking methods.
    \item \textbf{Special Diets} - Describes special diets.
\end{itemize}

For the rest of the pages, we scrape the \emph{Title}, the \emph{Ingredients} spearated by comma and the \emph{Instructions} as raw text separated by comma. The result is a dataset consisting of \textit{2204} rows and \textit{3} columns.
Furthermore, we check the existence of empty cells and we find that: \\

The column \emph{Title} has \textit{0} NA values. \\
The column \emph{Ingredients} has \textit{6} NA values. \\
The column \emph{Instructions} has \textit{630} NA values. \\

which we remove. We also notice that some instructions are not proper instructions so we need to filter them out. The histogram \ref{hc-t} below, showcases the length (in characters) of the instructions.

\begin{figure}[H]
    \centering
    \includegraphics[width=\textwidth,height=\textheight,keepaspectratio]{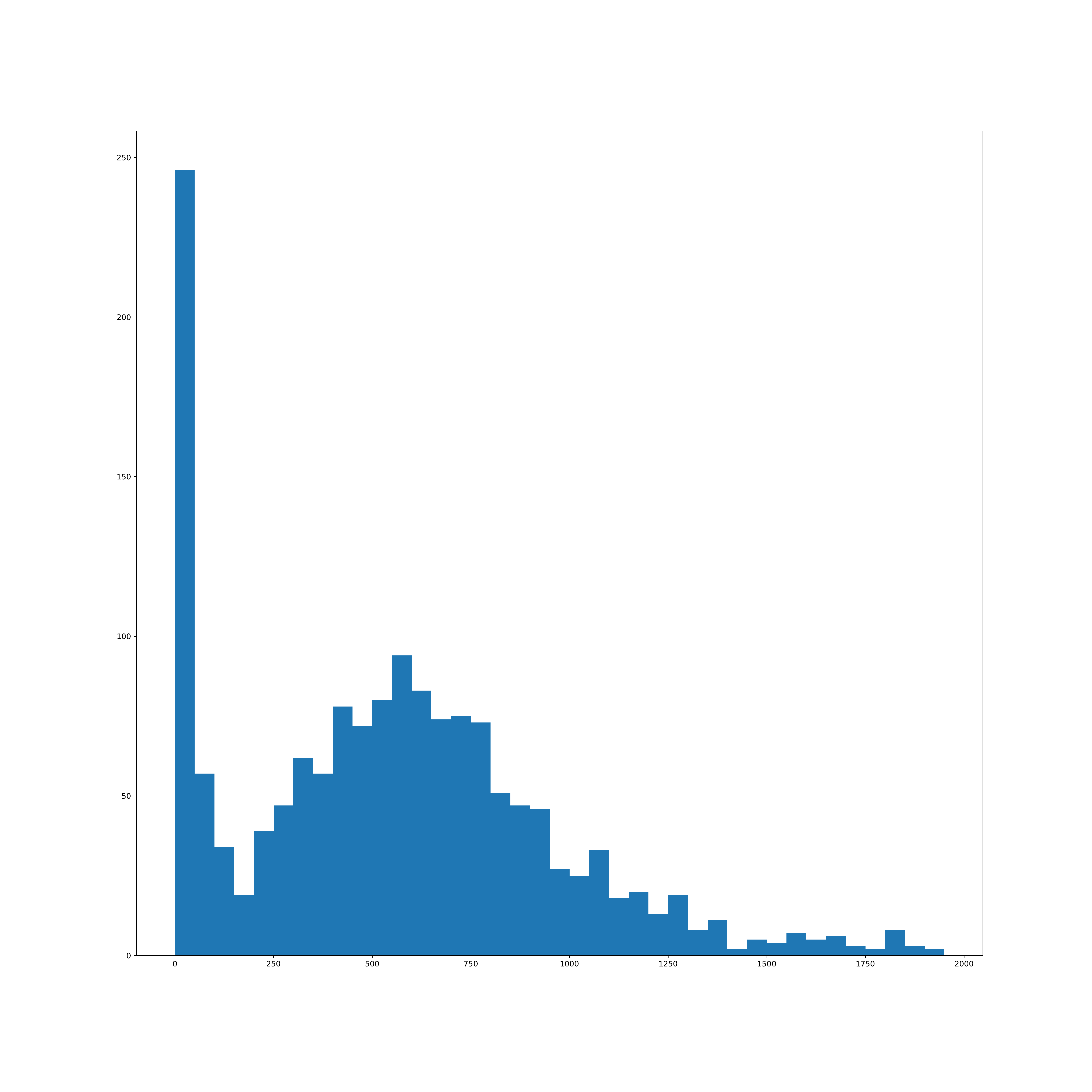}
    \caption{Histogram of character length of instructions}
    \label{hc-t}
\end{figure}

We see that the bulk of recipes has instructions of $\sim$0-50 characters which is not correct and must be unwanted text.
We investigate this below in the following Figure \ref{hc-mp}.

\begin{figure}[H]
    \centering
    \includegraphics[width=\textwidth,height=\textheight,keepaspectratio]{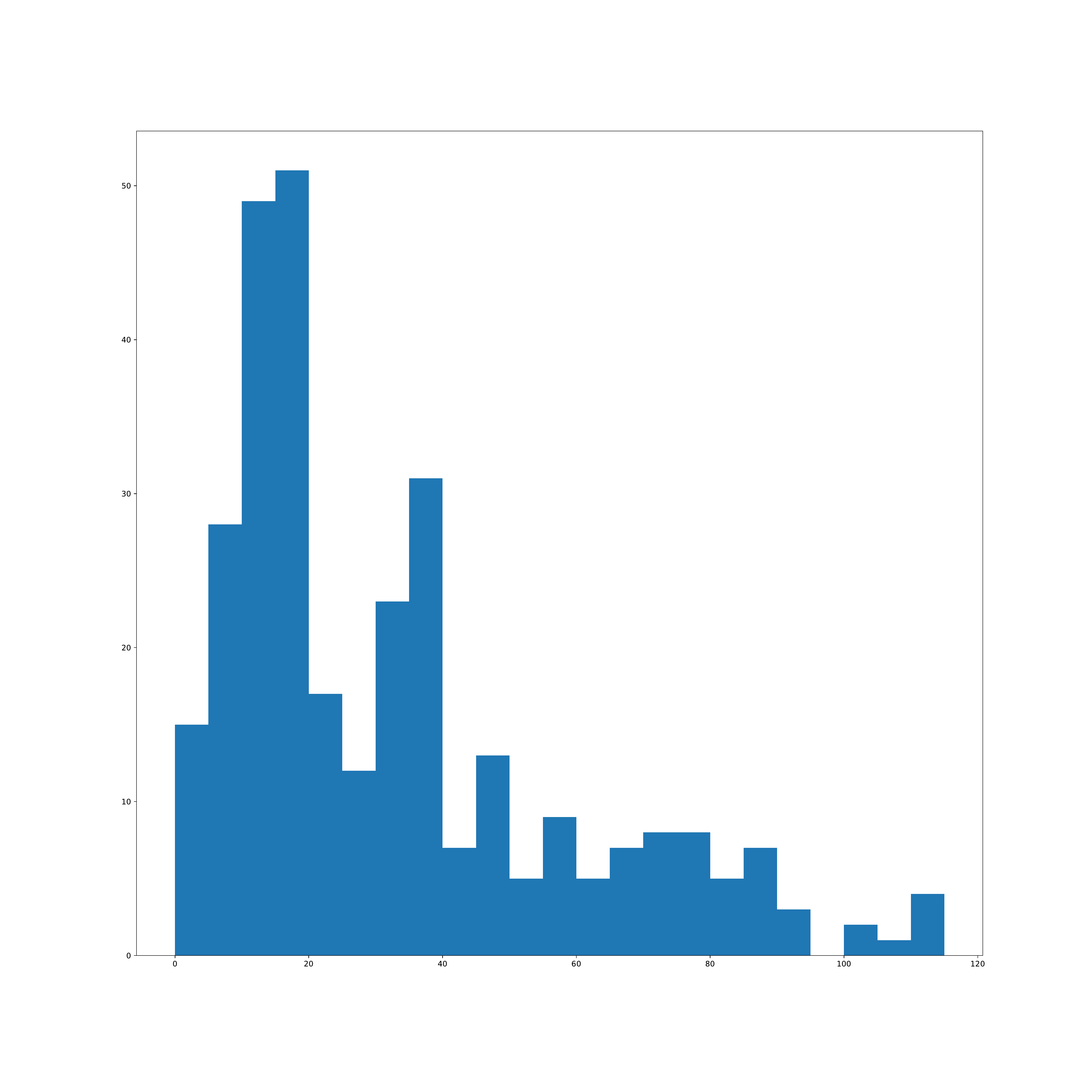}
    \caption{Histogram of character length of instructions (more precise scale)}
    \label{hc-mp}
\end{figure}

Indeed, the majority of the recipes has 0-50 characters. Furthermore, printing a sample of them reveals that indeed they are random / broken / wrong text:

\begin{verbatim}
     ,
    Wine pairing: Pomino Bianco,
    Wine pairing: Orvieto Classico,
    Dough
    ,
    Whoopie Pies
    Sour Butter
    For the beef tongue
    For the dried vegetable
\end{verbatim}

(Upon further inspection, we noticed that some instructions which are advertisement text pass the "over 50 character" filter and they all include the string: \textit{This recipe is taken from the book} and need to be removed) \\
We remove all of the above and following the same procedure we remove recipes without any ingredients. In the end, we have our final dataset called "data.csv" with \textit{1307} rows and \textit{3} columns which has the following information:

\begin{table}[h!]
\resizebox{\textwidth}{!}{ %
\centering
\begin{tabular}{|c | c | c|}
 \hline
 Title & Ingredients & Instructions\\
 \hline\hline
 Recipe title & Recipe ingredients. Comma separated & Recipe instructions. Comma separated\\
 \hline
\end{tabular}
}
\vspace{0.2cm}
\caption{Dataset field information}
\label{df-info}
\end{table}

\subsection{Data pre-processing}
Due to the nature of our problem, the dataset could be split into 2 columns but does not need to. Input text is what models usually will receive and target text is what we want them to generate. So the interaction looks like "Given the input text, we want the target text generated". Since what we are aiming to achieve is text generation, only the target texts can be used to fine-tune the model. Based on certain ingredients given, we need our input texts to be individual ingredients or combinations of them and have the model complete the recipe. A clever way of achieving this is using the ingredients of each recipe by taking the powerset of the ingredient list (the set of all subsets) and appending each element of that as the start of the target text. This way we manage to format our data in a way that is suitable for our task and we also managed to increase our dataset size as now the same recipe is the target for multiple ingredients. In the table \ref{df-split} below is the size comparison for our dataset.

\begin{table}[h!]
\centering
\begin{tabular}{|c | c |}
 \hline
 Before splitting & After splitting\\ [0.5ex]
 \hline\hline
 1307 & 11984\\
 \hline
\end{tabular}
\vspace{0.2cm}
\caption{Dataset split}
\label{df-split}
\end{table}

We see that we have a massive \textbf{816.9}\% increase in size which will be rather valuable. One last step is left until our dataset is ready for our task.

As with most text generation tasks, our text data will need some further pre-processing. To make it easier for the \emph{Transformers} to learn to identify important parts of the text / parts where there is a change in content. The ones that our project features are the title of the recipe, the ingredients and the instructions. We introduce certain \textit{tokens} which are string sequences that are nowhere present in our texts with the hope that the transformers understand that these arbitrary sequences are not part of the actual content, rather a sort of "meta-information" which conveys information about the contents. These sequences will be anything that is between a pair of \textbf{"< >"} (we decided on arrows, it could be any character that is not present within the content). In the end, the ones we need are:

\begin{itemize}
    \item \textbf{<START\_TITLE>} - Used to identify where the title starts
    \item \textbf{<START\_INGREDIENTS>} - Used to identify where the ingredients start
    \item \textbf{<START\_INSTRUCTIONS>} - Used to identify where the instructions start
\end{itemize}

On top of that (more as a cosmetic rather than a functional choice) we add the characters "\textbf{-}" and "\textbf{*}" as precursors to the ingredient and instruction lines respectively.

While the above choice might seem trivial, it enables us to represent the "target" text (the text we want our transformer to generate) as a simple string, a single sequence of text since the "separation" of the title, ingredients, and instructions happens semantically through the use of the aforementioned tokens. We also create a method that can decrypt the string and format it to a nice recipe structure and our dataset which now looks like this:

\begin{verbatim}
	lemons	<START_TITLE>Frog Meunière<START_INGREDIENTS>-...
	olive taste	<START_TITLE>Frog Meunière<START_INGREDIENTS>-...
	lemons	<START_TITLE>Frog Meunière<START_INGREDIENTS>-...
	bread	<START_TITLE>Frog Meunière<START_INGREDIENTS>-...
	frogs legs	<START_TITLE>Frog Meunière<START_INGREDIENTS>-...
\end{verbatim}

\section{Experiments}

The general idea we follow in our experiments is that of text generation. This task can be seen as the ability to predict the next character in a sequence and thus, character by character make the complete recipe. The model we are going to use is the \emph{GPT-2} which was trained with a \emph{CLM} (causal language modelling) objective in mind and can perform this task really well. As mentioned previously, we are going to access it through the "\textit{Hugging Face}" interface since it provides us with many methods for easier tuning and experimentation. This \href{https://huggingface.co/models?filter=gpt2}{Repository} features a plethora of fine-tuned GPT-2 models for a variety of tasks. We are going to create our own version however since none of these is trained on recipes. In the table \ref{diffs} below are all the available GPT-2 models.

\begin{table}[h!]
\centering
\begin{tabular}{|c | c | c | c| c|}
 \hline
 Version & Layers & Hidden & Heads & Parameters (In millions)\\ [0.5ex]
 \hline\hline
 gpt2 & 12 & 768 & 12 & 117\\
 gpt2-medium & 24 & 1024 & 16 & 345\\
 gpt2-large & 36 & 1280 & 20 & 774\\
 gpt2-xl & 48 & 1600 & 25 & 1558\\
 \hline
\end{tabular}
\vspace{0.2cm}
\caption{GPT-2 model differences}
\label{diffs}
\end{table}

\begin{figure}[H]
\centering
\includegraphics[scale=0.8]{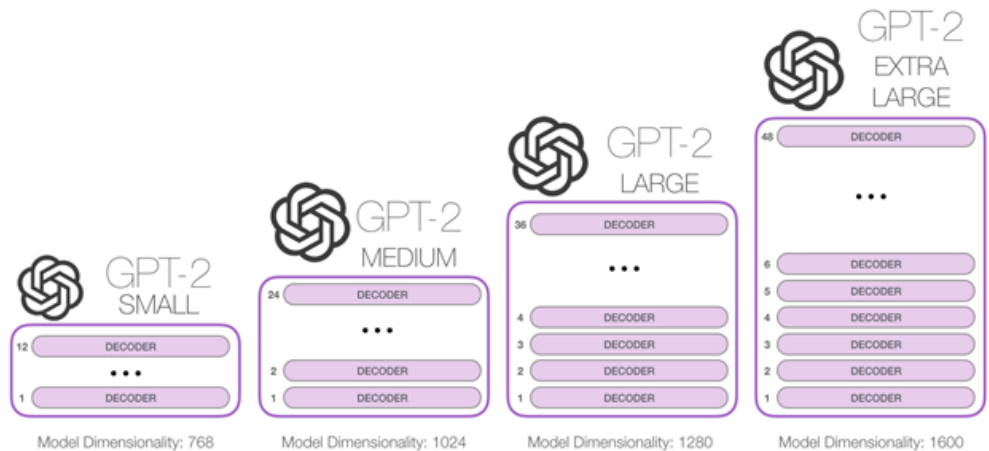}
\caption{GPT-2 model differences in architecture}
\label{diffs-graphical}
\end{figure}

Out of these we will using \emph{GPT2-Small} as for our task and dataset size a larger model would take a seriously longer time than we wanted without any benefit.
Within this GPT-2 "framework" are the following useful classes we are going to use \textit{GPT2LMHeadModel,  GPT2Tokenizer, GPT2Config}. Each of them is going to be explained in the following section along how it was used.

\subsection{Setup}

The setup for our project was relatively straightforward. By following a structured way of setting up the following classes, we were able to get the model up and going for training fast, thanks to the easy and efficient interface of Hugging Face. The very first thing we needed to is create a dictionary of tokens. These tokens are some used by the model, and some that we have created. At this point it is extremely important to note that any "meta" token that introduced in our text must be made know to the tokenizer. Thus, the following dictionary was made:

\begin{Verbatim}[breaklines=true]
    {
        "bos_token": "<|startoftext|>",
        "eos_token": "<|endoftext|>",
        "pad_token": "<|pad|>",
        "additional_special_tokens": ["<START_TITLE>", "<START_INGREDIENTS>", "<START_INSTRUCTIONS>"]
    }
\end{Verbatim}

It is clear that the first 3 entries, are tokens required by the model and thus we include them as is. Note that the tokens themselves are required, not the exact text, so in another project the "\textit{bos\_token}" could have the value "\textit{<bos\_text>}" instead. It is only important that it exists. Obviously the first and second token indicate the beginning and end of the text respectively and the "\textit{pad\_token}" is used for padding all the recipes to the same length. The last entry consists of a list of our tokens which must be made know to the tokenizer so we include all of them, the same way they appear in the texts.

\subsubsection{GPT2Tokenizer}

The \emph{GPT2Tokenizer} class, is used to construct a gpt tokenizer. As beautifully explained in this \href{https://lukesalamone.github.io/posts/gpt2-tokenization/}{article} by Luke Salamone, most deep learning models cannot (and in fact should not) work directly with strings. Instead, strings are broken down into pieces, tokens. Each token can be whatever works best for the task at hand be it a word, a phrase or even a character, and then they are turned into (represented by) numbers (which is what computers can work with). As an example of this process, we will use the \emph{GPT2Tokenizer} to see how the phrase "I love food" gets turned into numbers.

\begin{Verbatim}
    {'input_ids': [40, 1842, 2057], 'attention_mask': [1, 1, 1]}
\end{Verbatim}

Every word of that sentence was turned into a number as visible in the \emph{input\_ids} list (each integer in the list corresponds to a word in the string). The decoding process does the reverse and is able to go from integers back into strings. This entire process can only happen if a dictionary exists. In the training phase, the tokenizer parses are tokens and all the integers are accumulated and they create the vocabulary which is the entirety of the train corpus as integers. Thus, we understand that for different vocabularies, the same word might be represented by a different integer. Below is a table \ref{tm-size} showcasing the different vocabularies of different transformers \cite{TransformerTokenizers}:

\begin{table}[h!]
\centering
\begin{tabular}{|c | c |}
 \hline
 Transformer model & Vocabulary size\\ [0.5ex]
 \hline\hline
    bert-base-uncased & 30522 \\
    bert-base-cased & 28996 \\
    bert-base-multilingual-cased & 119547 \\
    xlm-mlm-en-2048 & 30145 \\
    gpt2 \& roberta & 50257 \\
    word2vec & 3000000 \\
    glove & 400000 \\
 \hline
\end{tabular}
\vspace{0.2cm}
\caption{Transformer models and vocabulary size}
\label{tm-size}
\end{table}

A few other methods used frequently in conjunction with tokenization to configure vocabulary size and/or quality are these of \emph{Stemming} and \emph{Lemmatization}. Stemming essentially "cuts" the suffixes of words, reducing them to their root forms (stems).

\begin{figure}[H]
\centering
    \begin{tikzpicture}
    \node[font=\Large] (first) at (-3,8) {connected};
    \node[font=\Large] (second) at (-3,7.2) {connection};
    \node[font=\Large] (third) at (-3,6.4) {connector};
    \node[font=\Large] (fourth) at (-3,5.6) {connectivity};
    \node[font=\Large] (fifth) at (2,6.8) {connect};
    \draw[->, very thick] (first) edge (fifth);
    \draw[->, very thick] (second) to (fifth);
    \draw[->, very thick] (third) to (fifth);
    \draw[->, very thick] (fourth) edge (fifth);
    \end{tikzpicture}
\caption{Stemming example}
\end{figure}
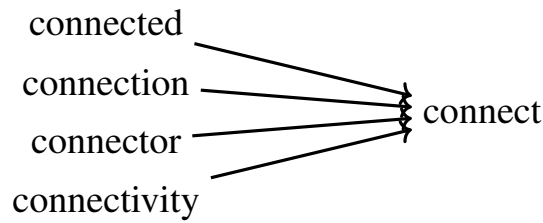

Lemmatization is the process of turning words back into the base of the word, the \textit{lemma}, through the use of a vocabulary and analysis of the word and is a more advanced method than stemming.

\begin{figure}[H]
\centering
    \begin{tikzpicture}
    \node[font=\Large] (first) at (-3,8) {player};
    \node[font=\Large] (second) at (-3,7.2) {am, are, is};
    \node[font=\Large] (third) at (-3,6.4) {studying};
    \node[font=\Large] (fourth) at (-3,5.6) {connectivity};
    \node[font=\Large] (fifth) at (2,8) {play};
    \node[font=\Large] (sixth) at (2,7.2) {be};
    \node[font=\Large] (seventh) at (2,6.4) {study};
    \node[font=\Large] (eighth) at (2,5.6) {connect};
    \draw[->, very thick] (first) edge (fifth);
    \draw[->, very thick] (second) to (sixth);
    \draw[->, very thick] (third) to (seventh);
    \draw[->, very thick] (fourth) edge (eighth);
    \end{tikzpicture}
\caption{Lemmatization examples.}
\end{figure}

The entire process, usually takes the form shown below even though variations of it can also happen.

\begin{figure}[H]
\resizebox{\textwidth}{!}{
    \begin{tikzpicture}[
        squarednode/.style={rectangle, draw=blue!60, fill=blue!5, very thick, minimum size=10mm},
    ]
        \node[squarednode] (start) {Normalization};
        \node[squarednode] (sec) [right=of start] {Tokenization};
        \node[squarednode] (thi) [right=of sec] {Stopword removal};
        \node[squarednode] (fou) [right=of thi] {Stem / Lemmatize};

        \draw[->] (start.east) -- (sec.west);
        \draw[->] (sec.east) -- (thi.west);
        \draw[->] (thi.east) -- (fou.west);
    \end{tikzpicture}
}
\caption{NLP pre-processing pipeline.}
\end{figure}
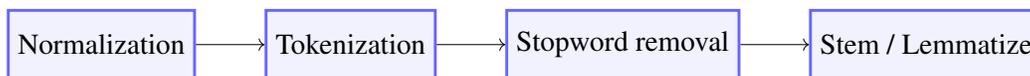

However, the \textit{GPT2Tokenizer} uses a different method called Byte Pair Encoding (BPE). This algorithm starts with tokens that are two bytes in size and based on the frequency of appearance of their pairs, it concatenates them into a larger token until we reach the wanted vocabulary size. For example, if the letters "h" and "e" appear together a lot of the time, it concatenates them into the token "he", if the tokens "s" and "he" also appear frequently together they get concatenated to "she" and so on.

All of the above, come "packaged" together with the pre-trained GPT2Tokenizer so all we have to do is import the pre-trained class. Using the \textit{.add\_special\_tokens()} method we pass the extra tokens that we have added to our texts. This is the first instance of "fine tuning" we do since the pre-trained model now received a list of different tokens specialized for our purpose. This tokenizer treats spaces as parts of the tokens so a word is differently encoded whether it is on the begging, or the end of the sequence (or if it has no whitespace).

Last but not least, we need to explain \emph{Attention Masks}. If you remember in the example above, when we tokenized the phrase "I love food" we got a list named "attention\_mask" along with the integers. Attention masks are tensors that are used when we need to perform inference fast so they are not 100\% essential (even though they help a lot). The difference between slow and fast inference is batching. Models, cannot work with tensors of different sizes so sequences of different length are padded to the same length (remember the "\textit{pad\_token}"). Attention masks are essentially masks that help the tokenizers understand which tokens are purely for padding purposes and which ones are content, having 0s where the pad tokens are and 1s where the content ones are.

\subsubsection{GPT2Config}
The GPT-2's model configuration is stored in this configuration class. A GPT-2 model is instantiated using the specified arguments, defining the model architecture. An instantiated configuration that uses the default settings will have a similar setup to the GPT-2 gpt2 architecture. The (most important) default arguments as outlined in the \emph{Hugging Face} documentation are showed in the following Table \ref{arguments}.

\begin{table}[h!]
\centering
\begin{tabular}{|c|c|p{7cm}|}
 \hline
 Argument & Default value & Explanation\\ [0.5ex]
 \hline\hline
    input\_ids & 50257 & This is the vocabulary size of the model. Defines the amount of different tokens that can be represented by the inputs\_ids \\
    n\_positions & 1024 & The maximum length of a sequence that the model might ever meet \\
    n\_ctx & 1024 &  Dimensionality of the causal mask (usually same as n\_positions)\\
    n\_embd & 768 & Dimensionality of the embeddings and hidden states \\
    n\_layer & 12 & Number of hidden layers in the Transformer encoder\\
    n\_head & 12 &  Number of attention heads for each attention layer in the Transformer encoder\\
    activation\_function & 'gelu' & Activation function\\
    resid\_pdrop & 0.1 & The dropout probability for all fully connected layers in the embeddings, encoder, and pooler\\
    embd\_pdrop  & 0.1 & The dropout ratio for the embeddings\\
    attn\_pdrop & 0.1 & The dropout ratio for the attention\\
 \hline
\end{tabular}
\vspace{0.2cm}
\caption{GPT2Config parameters and default values.}
\label{arguments}
\end{table}

We proceed with these default values and use the class to instantiate a model configuration used in conjuction with the \emph{GPT2LMHeadModel} class.

\subsubsection{GPT2LMHeadModel}
This is the actual model class. The reason behind its name is the fact that this GPT-2 model includes a language modeling head on top of the existing transformer. Basically a linear layer connected to the input embeddings which uses the d-dimensional representation from the transformer to predict what the next token in the sequence is. The main difference between the \emph{GPT2LMHeadModel} and \emph{GPT2} classes is the fact that the \emph{GPT2} class does not include any head on top of the transformer and thus simply outputs raw hidden-states. The only parameter is the aforementioned configuration object. So the model is almost ready.

Using the method \textit{resize\_token\_embeddings()} and passing it the number of new tokens (tokens we have introduced ourselves). It's main job is to add newly initialized vectors at the end of the embedding matrix when we have a token embedding size increase and remove vectors from the end when we have a decrease.

\begin{center}
    \begin{tikzpicture}[every node/.style={draw, minimum size=1cm}]
        \matrix[draw=none] (mymat)
          {
            \node {1.2}; & \node{-0.1}; & \node {4.3}; & \node {3.2}; \\
            \node {0.4}; & \node{2.5}; & \node {-0.9}; & \node {0.5}; \\
            \node {2.1}; & \node{0.3}; & \node {0.1}; & \node {0.4}; \\
          };
          \matrix[draw=none, right= 1.5cm of mymat.east] (mysecmat)
          {
            \node (a) {0.2}; \\
            \node (b) {-1.4}; \\
            \node (c) {2.1}; \\
          };
          \node[draw=none, above= 17mm of mymat.east, font=\Large, font=\bf, xshift=-2em] {Token embedding increase};
          \draw[->, thick] (mysecmat.west) to (mymat.east);
          \draw[very thick,red] (c.south west) |- (a.north east);
          \draw[very thick,red] (c.south west) -| (a.north east);
    \end{tikzpicture}

    \begin{tikzpicture}[every node/.style={draw, minimum size=1cm}]
        \matrix[draw=none] (mysthirmat)
          {
            \node {1.2}; & \node{-0.1}; & \node {4.3}; & \node {3.2}; & \node (aa) {0.2};\\
            \node {0.4}; & \node{2.5}; & \node {-0.9}; & \node {0.5}; & \node {-1.4};\\
            \node {2.1}; & \node{0.3}; & \node {0.1}; & \node {0.4}; & \node (cc) {2.1};\\
          };
          \node[draw=none, above=1mm of mymat, font=\Large, font=\bf] {Token embedding decrease};
          \node[draw=none, below= of mysthirmat, yshift=3em] {Embedding matrix};
          \draw[very thick,red] (cc.south west) |- (aa.north east);
          \draw[very thick,red] (cc.south west) -| (aa.north east);
          \draw[very thick] (cc.south west) -- (aa.north east);
          \draw[very thick] (cc.south east) -- (aa.north west);
    \end{tikzpicture}
\end{center}

\subsubsection{GPT2Dataset}

On top of all this, we created a custom dataset for our model. The class \textit{GPT2Dataset} represents this custom dataset and has the following attributes:

\begin{itemize}
    \item \textbf{tokenizer} - This is the tokenizer class we instantiated previously and is used to encode (and later decode) the texts.
    \item \textbf{input\_ids} - This is a list of the IDs of the inputs
    \item \textbf{attn\_masks} - This is a list of attention masks
\end{itemize}

As well as the methods \textit{\_\_len\_\_()} which returns the size of the dataset and \textit{\_\_getitem\_\_()} which returns an (ID, Attention mask) pair.
The constructor for our class takes the following parameters:

\begin{itemize}
    \item \textbf{txt\_list} - This is the list of text we pass to the model (our \emph{Data}).
    \item \textbf{tokenizer} - The tokenizer we mentioned previously.
    \item \textbf{gpt2\_type} - The type of GPT model we are using. (it is the "\textit{gpt2}").
    \item \textbf{max\_length} - Max length of any of the texts we pass (we default it to 1000 characters).
\end{itemize}

The final task with regards to data happens here and is just appending the tokens \textit{<|startoftext|>} and \textit{<|endoftext|>} before and after each text we read into our dataset. With that, we shuffle the data and prepare the dataset by creating an object and passing it to a \emph{Pytorch} dataloader class. This class takes care of shuffling and batching for us, and generally makes passing data through the model much easier.

The split we go for is a 80-20\% train-test split and a 90-10\% train-validation split. That leaves us with \textbf{8628} training samples, \textbf{959} validation samples and \textbf{2397} test samples.
At this point, we need to check whether the split, is actually stratified. Meaning we need to check whether the training ans test list have the same distributions of letters and first letters. If there is a large percentage difference of letters this could bias the model and/or overfit it towards certain characters. Firstly we are gonna check the distribution of the first characters of each sequence and then the total characters present in each sequence.

Regarding the first characters of each sequence, below is the table \ref{chars} for percentage distribution of first characters as percentages of the total first characters.

\begin{table}[h!]
\centering
\begin{tabular}{lrrr}
\toprule
Character & Train list (\%)& Test list (\%)&  Difference \\
\midrule
  a &  1.935337 &  2.231330 &               0.295993 \\
  b &  6.306922 &  6.284153 &               0.022769 \\
  c & 14.264572 & 13.433515 &               0.831056 \\
  d &  3.324226 &  2.914390 &               0.409836 \\
  e &  1.969490 &  1.912568 &               0.056922 \\
  f &  7.160747 &  6.466302 &               0.694444 \\
  g &  4.895264 &  5.373406 &               0.478142 \\
  h &  1.707650 &  1.411658 &               0.295993 \\
  i &  0.182149 &  0.045537 &               0.136612 \\
  j &  0.990437 &  1.229508 &               0.239071 \\
  k &  0.318761 &  0.591985 &               0.273224 \\
  l &  4.508197 &  5.828780 &               1.320583 \\
  m &  5.202641 &  4.872495 &               0.330146 \\
  n &  1.047359 &  1.229508 &               0.182149 \\
  o &  5.122951 &  5.373406 &               0.250455 \\
  p &  7.627505 &  7.559199 &               0.068306 \\
  q &  0.159381 &  0.273224 &               0.113843 \\
  r &  2.777778 &  2.914390 &               0.136612 \\
  s & 15.767304 & 15.346084 &               0.421220 \\
  t &  5.965392 &  5.919854 &               0.045537 \\
  u &  0.512295 &  0.591985 &               0.079690 \\
  v &  3.711293 &  3.961749 &               0.250455 \\
  w &  2.959927 &  2.777778 &               0.182149 \\
  x &  0.250455 &  0.136612 &               0.113843 \\
  y &  1.218124 &  1.138434 &               0.079690 \\
  z &  0.113843 &  0.182149 &               0.068306 \\
\bottomrule
\end{tabular}
\vspace{0.2cm}
\caption{First character distributions for train and testing lists}
\label{chars}
\end{table}

As we can see, there does not seem to be any significant difference in the distribution of the first letters.

Also, the total distribution of letters in sequences does not seem to show any important percentage differences.

\begin{table}[h!]
\begin{center}
\begin{tabular}{lrrr}
\toprule
Character & Train list (\%) & Test list (\%) &  Difference \\
\midrule
  a &  8.162175 &  8.155947 &               0.006228 \\
  b &  1.712851 &  1.729638 &               0.016787 \\
  c &  3.630478 &  3.646519 &               0.016041 \\
  d &  3.998473 &  3.995703 &               0.002769 \\
  e & 11.815909 & 11.859382 &               0.043473 \\
  f &  1.804570 &  1.801777 &               0.002793 \\
  g &  2.873736 &  2.845883 &               0.027854 \\
  h &  4.369251 &  4.376343 &               0.007092 \\
  i &  6.906201 &  6.885347 &               0.020854 \\
  j &  0.140189 &  0.142740 &               0.002551 \\
  k &  1.077238 &  1.071158 &               0.006081 \\
  l &  4.977452 &  4.987402 &               0.009950 \\
  m &  2.349569 &  2.365257 &               0.015687 \\
  n &  6.754983 &  6.757937 &               0.002954 \\
  o &  6.720791 &  6.677947 &               0.042844 \\
  p &  3.186295 &  3.216550 &               0.030255 \\
  q &  0.084367 &  0.090651 &               0.006285 \\
  r &  6.538899 &  6.524068 &               0.014830 \\
  s &  6.509116 &  6.509800 &               0.000684 \\
  t &  9.608021 &  9.588437 &               0.019584 \\
  u &  3.010516 &  3.014083 &               0.003567 \\
  v &  0.971201 &  0.971224 &               0.000023 \\
  w &  1.338396 &  1.319454 &               0.018943 \\
  x &  0.278953 &  0.274075 &               0.004878 \\
  y &  1.001823 &  1.000769 &               0.001053 \\
  z &  0.144371 &  0.153720 &               0.009348 \\
  º &  0.009270 &  0.009813 &               0.000543 \\
  à &  0.000506 &  0.000637 &               0.000130 \\
  â &  0.000639 &  0.001114 &               0.000475 \\
  ç &  0.000666 &  0.000637 &               0.000029 \\
  è &  0.003143 &  0.002970 &               0.000173 \\
  é &  0.014612 &  0.016178 &               0.001567 \\
  ê &  0.000639 &  0.000849 &               0.000209 \\
  î &  0.000999 &  0.001220 &               0.000221 \\
  ñ &  0.001785 &  0.002387 &               0.000602 \\
  ò &  0.000639 &  0.000849 &               0.000209 \\
  ø &  0.000186 &  0.000106 &               0.000080 \\
  ù &  0.000133 &  0.000159 &               0.000026 \\
  ú &  0.000639 &  0.000849 &               0.000209 \\
  $\acute{\hat{\text{e}}}$ &  0.000320 &  0.000424 & 0.000105 \\
\bottomrule
\end{tabular}
\vspace{0.2cm}
\caption{Total character distributions for train and testing lists}
\label{chars-total}
\end{center}
\end{table}

\subsection{Training}

For the training, we set the following parameters.

\textbf{epochs} = 3

\textbf{learning\_rate} = 5e-4 (0.0005)

\textbf{warmup\_steps} = 1e2 (100)

\textbf{epsilon} = 1e-8 (0.00000001)

\textbf{Epochs} are the passes over our complete dataset. One epoch equals a complete pass over our entire dataset, meaning our entire dataset has been through the model.

\textbf{Learning rate} is a tuning parameter which essentially determines how large the step we take towards the minimum of the loss function is when updating the models' weights. The value of the learning rate is important because it directly impacts the learning process.

Very large learning rates can cause the step to "overshoot" the minimum and get stuck in an loop of jumping around or even make the loss worse as can be seen in the figure \ref{llr} below.

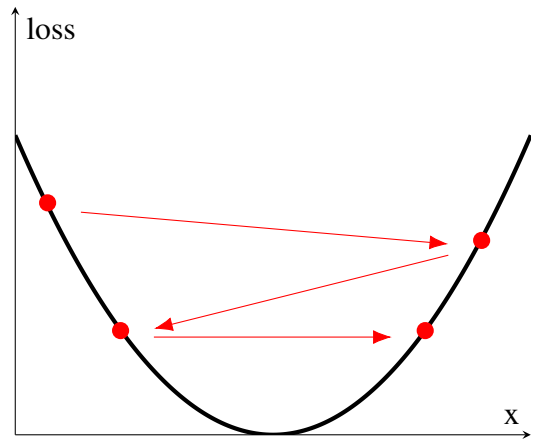
\begin{figure}[H]
    \centering
    \begin{tikzpicture}
        \pgfplotsset{ticks=none}
        \begin{axis}[axis lines=middle, tick style={very thick},xlabel={x}, ylabel={loss},]
            %
            \addplot[yscale=0.7,ultra thick,samples=151,domain=0:10] {0.3*(x-5)^(2) + 1};
        \end{axis}
        \node[scale=2, red] (1) at (0.43,3) {\textbullet};
        \node[scale=2, red] (2) at (6.2,2.5) {\textbullet};
        \node[scale=2, red] (3) at (1.4,1.3) {\textbullet};
        \node[scale=2, red] (4) at (5.45,1.3) {\textbullet};

        \draw[-{Latex[length=2.5mm]}, red] (1) -- (2);
        \draw[-{Latex[length=2.5mm]}, red] (2) -- (3);
        \draw[-{Latex[length=2.5mm]}, red] (3) -- (4);
        \newline
    \end{tikzpicture}
\vspace{0.2cm}
\caption{Large learning rate}
\label{llr}
\end{figure}

Very small learning rates cause the model to take really small steps towards the loss function's minimum and while not an issue, this can cause the model to have a really long optimization process which slows our model down for no reason as shown in the figure \ref{slr} below.

\begin{figure}[H]
    \centering
    \begin{tikzpicture}
        \pgfplotsset{ticks=none}
        \begin{axis}[axis lines=middle, tick style={very thick},xlabel={x}, ylabel={loss},]
            %
            \addplot[yscale=0.7,ultra thick,samples=151,domain=0:10] {0.3*(x-5)^(2) + 1};
        \end{axis}
        \node[scale=2, red] (1) at (0.43,3) {\textbullet};
        \node[scale=2, red] (2) at (0.55,2.7) {\textbullet};
        \node[scale=2, red] (3) at (0.71,2.4) {\textbullet};
        \node[scale=2, red] (4) at (0.87,2.1) {\textbullet};
        \node[scale=2, red] (5) at (1.05,1.8) {\textbullet};
        \node[scale=2, red] (6) at (1.27,1.5) {\textbullet};

        \node[transparent] (6) at (0.15,3) {\textbullet};
        \node[transparent] (7) at (0.3,2.7) {\textbullet};
        \node[transparent] (8) at (0.51,2.4) {\textbullet};
        \node[transparent] (9) at (0.67,2.1) {\textbullet};
        \node[transparent] (10) at (0.9,1.8) {\textbullet};
        \node[transparent] (11) at (1.2,1.5) {\textbullet};

        \draw [<-] (6) -- (7);
        \draw [<-] (7) -- (8);
        \draw [<-] (8) -- (9);
        \draw [<-] (9) -- (10);
        \draw [<-] (10) -- (11);
        \newline
    \end{tikzpicture}
\vspace{0.2cm}
\caption{Small learning rate}
\label{slr}
\end{figure}
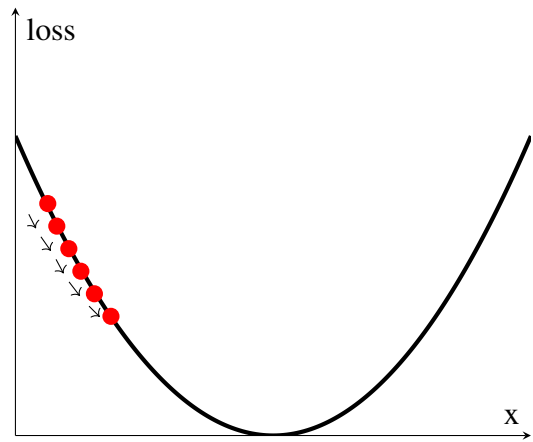

With the optimal learning rate we reach the (possibly local) minimum in consistent and fast enough steps. Thus, in order to figure out the optimal value trying various learning rates is required. Some values that are used regularly are 0.1, 0.01, 0.001.

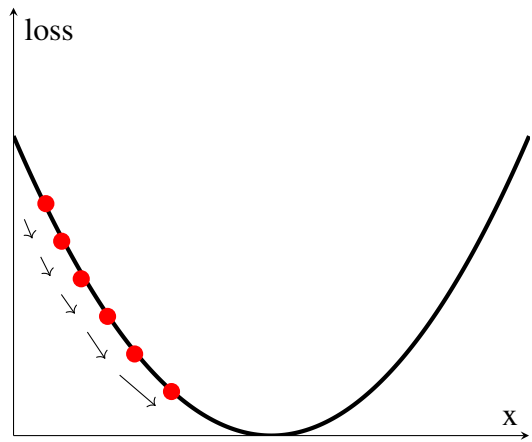
\begin{figure}[H]
    \centering
    \begin{tikzpicture}
        \pgfplotsset{ticks=none}
        \begin{axis}[axis lines=middle, tick style={very thick},xlabel={x}, ylabel={loss},]
            %
            \addplot[yscale=0.7,ultra thick,samples=151,domain=0:10] {0.3*(x-5)^(2) + 1};
        \end{axis}
        \node[scale=2, red] (1) at (0.43,3) {\textbullet};
        \node[scale=2, red] (2) at (0.64,2.5) {\textbullet};
        \node[scale=2, red] (3) at (0.9,2) {\textbullet};
        \node[scale=2, red] (4) at (1.25,1.5) {\textbullet};
        \node[scale=2, red] (5) at (1.61,1) {\textbullet};
        \node[scale=2, red] (6) at (2.1,0.5) {\textbullet};

        \node[transparent, scale=0.5] (60) at (0.093,3) {\textbullet};
        \node[transparent, scale=0.5] (70) at (0.3,2.5) {\textbullet};
        \node[transparent, scale=0.5] (80) at (0.55,2) {\textbullet};
        \node[transparent, scale=0.5] (90) at (0.9,1.5) {\textbullet};
        \node[transparent, scale=0.5] (100) at (1.3,0.9) {\textbullet};
        \node[transparent, scale=0.5] (110) at (2,0.3) {\textbullet};

        \draw [->] (60) -- (70);
        \draw [->] (70) -- (80);
        \draw [->] (80) -- (90);
        \draw [->] (90) -- (100);
        \draw [->] (100) -- (110);
        \newline
    \end{tikzpicture}
\vspace{0.2cm}
\caption{Optimal learning rate.}
\end{figure}

For this reason, we use learning rate schedulers (and within them optimizers) and one of two scheduling techniques between \textit{Constant learning rate}, which as the name suggests is the strategy of choosing a learning rate and leaving it unchanged for the entire training process and \textit{Learning rate decay} where an initial value for is chosen but it is slowly reduced (thus the "decay") according to the scheduler.
This is done as the value of the learning rate needs to change between epochs during training to reflect the fact that early in the training process, the change of the model's weight will need a much larger change rather than the smaller, finer changes that are required later on.
The most common schedules are \textit{time-based decay} which reduces the learning rate in accordance to this formula $lr = lr_0/(1+k_t)$ with \textbf{k} being a hyperparameter and \textbf{t} being the iteration number, \textit{step decay} which reduces learning rate by some factor every few epochs with that number being a hyperparameter and \textit{exponential decay} which reduces the learning rate in accordance to this formula $lr = lr_0*e^{(-k_t)}$ with \textbf{k} being a hyperparameter and \textbf{t} being the iteration number.

Another technique that can be used is that of \textit{Adaptive learning rate methods} which is a group of methods that aim to achieve the same goal with the scheduler with the added benefit that they require no hyperparameter tuning as everything is done heuristically.
Some of these algorithms are \textit{Adagrad}, \textit{RMSprop} and \textit{Adam}.

Out of the more "modern" optimizers, the we tried the following: \textit{Adam}, \textit{AdamW} and \textit{Amsgrad}. However due to the nature of our data (small batch size as well as dataset size) and more importantly the results of this article by \emph{FastAI} \citet{fastAi} on why \textit{AdamW} produces the best results, we decided to stick with it.

\textbf{Warmup steps} are simply the amount of steps (as in training steps) during which our learning rate will remain unchanged or will even increase. After the steps have passed, the scheduler takes over and starts the decay. This is used since we are using an optimizer (Adam) which needs to calculate certain statistics for the gradients.

Having set everything up we calculate the steps per epoch as $len(dataloader) * epochs$ which comes out as 12942

For the training loop, we will execute the following steps in order:

\begin{enumerate}
    \item Firstly we request the next batch of data from our dataloader (see previous section about the dataset) which is pretty easy while using the \href{https://pytorch.org/docs/stable/data.html#torch.utils.data.DataLoader}{dataloader} class of \textit{Pytorch}. After retrieving the next batch, we split it into input IDs, labels and attention masks and move all of them to the GPU for faster computations using the "\emph{.to()}" method and passing the 'cuda' (GPU) device to it.
    \item The next step is to zero out the gradients of the model before every "backwards" step since by default the gradients that are computed will be accumulated instead of being replaced. Meaning that gradients won't change variables and will instead be summed over the course of several steps before they cumulatively affect variables. That works well witn \emph{RNNs} butin our case we will be zeroing them after every backwards step.
    \item After that, we calculate the model's outputs based on the batch we have by calling our model and passing to it the triplet we get from our batch. It is important that we move the model on the same device (for us this is the '\textit{cuda}' device, the GPU) as the data, but that needs to happen only once so we have done that before the training loop.
    \item This is possibly the most important step in the entire training loop because it is the step that enables the model to actually do the learning. It is where we (or rather have the model itself) calculate the loss based on our outputs of the previous step. This loss is stored in an array, and depending on the step we are on, it gets logged to the output.
    \item Then, by calling \emph{loss.backward()}, the $\frac{\partial loss}{\partial x}$ is computed for every parameter that requires gradients to be computed (i.e. has \textit{requires\_grad} = True) and are then accumulated in the \emph{x.grad} member of every parameter.
    \item Finally, using the \emph{optimizer.step()} method, x values are updated according to the aforementioned \textit{x.grad} gradient value of each parameter. \emph{Scheduler.step()} is then used to update the learning rate, as discussed in the previous chapters about learning rates.
\end{enumerate}

This is the entirety of the training loop which happens until we loop through the entire dataset (one epoch as we explained). After each training epoch (3 in our case), certain statistics are computed and logged. We calculate the average training loss as $\frac{\text{total training loss}}{\text{dataset length}}$ as well as the train perplexity. Perplexity is a common (if not the most common) metric for autoregressive language model evaluation like the \emph{GPT-2} we have used. Before we explain perplexity however, it is important to have an overview of basic language model terminology and metrics.

Given any language \emph{L}, most models calculate probabilities for symbol sequences say $(w_1, w_2, w_3, ... , w_n)$ according to their probability of existence in the language as such:

\begin{center}
$P(w_1, w_2, w_3, ... , w_n) = p(w_1)p(w_2|w_1)p(w_3|w_1,w_2)...p(w_n|w_1,w_2,w_3,...,w_\text{n-1}) =  \prod_{i=1}^{n}p(w_i|w1,w_2,w_3,...,w_\text{i-1})$
\end{center}

where $w_1,w_2,w_3,...,w_n$ can be anything from a single character, a word or a sub-word. So for example, for the language $\emph{L} = English \cap \{"the", "car", "crashed", "blue", "yellow", "run"\}$, the probability of the sentence "the blue car crashed" is computed as follows:

\begin{center}
$P("\text{the blue car crashed}") = p(``the``)p("blue"|"the")p("car"|"the","blue")p("crashed"|"the","blue","car")$
\end{center}
so below, is the model's distribution of probabilities of each word, based on how likely it is to be the first word of the sentence.

\begin{figure}[H]
\centering
\begin{tikzpicture}
\begin{axis}[
     ybar,
     ytick=\empty,
     axis y line=left,
     ylabel near ticks,
     width  = 12cm,
     axis x line*=bottom,
     height = 8cm,
     bar width=20pt,
     symbolic x coords={the, car, crashed, blue, yellow, run},
     nodes near coords,
     ymin=0,
     ylabel={$P(w_1)$}
     ]
     \addplot[ybar, fill=blue!60] coordinates {
          (the,0.4)
          (car,0.2)
          (crashed,0.15)
          (blue,0.1)
          (yellow,0.1)
          (run,0.05)
     };
\end{axis}
\end{tikzpicture}
\caption{First word probability distribution.}
\end{figure}
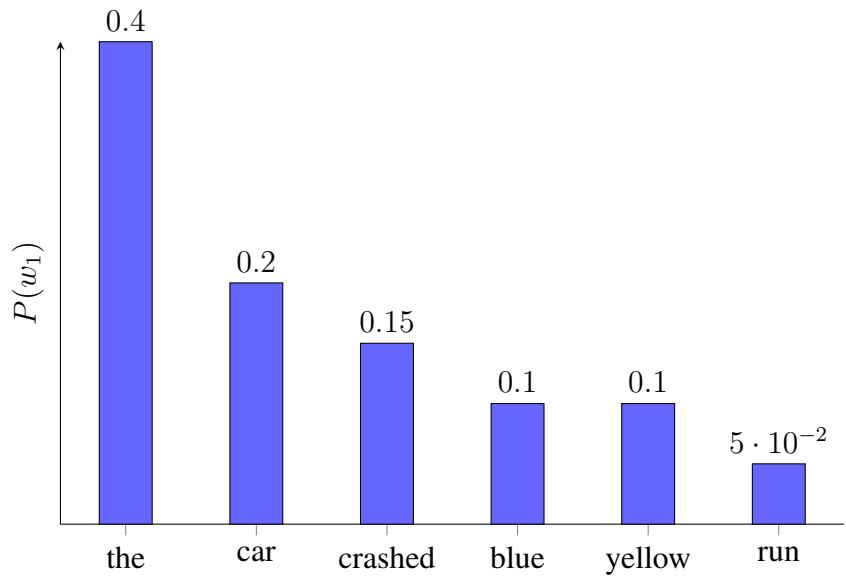

Having chosen "the" as the first word, we also have the model's distribution of probabilities of each word, based on how likely it is to be the next word in the sentence.

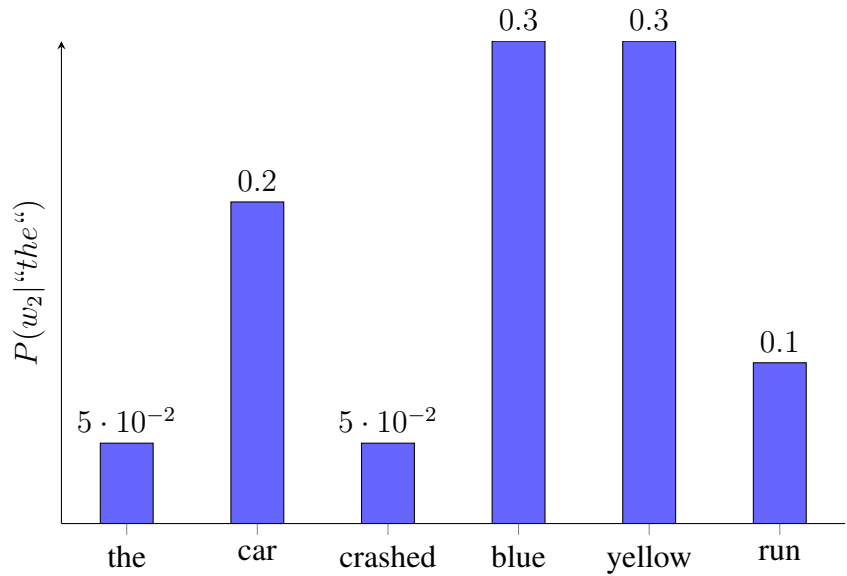
\begin{figure}[H]
\centering
\begin{tikzpicture}
\begin{axis}[
     ybar,
     ytick=\empty,
     axis y line=left,
     ylabel near ticks,
     width  = 12cm,
     axis x line*=bottom,
     height = 8cm,
     bar width=20pt,
     symbolic x coords={the, car, crashed, blue, yellow, run},
     nodes near coords,
     ymin=0,
     ylabel={$P(w_2|``the``)$}
     ]
     \addplot[ybar, fill=blue!60] coordinates {
          (the,0.05)
          (car,0.2)
          (crashed,0.05)
          (blue,0.3)
          (yellow,0.3)
          (run,0.1)
     };
\end{axis}
\end{tikzpicture}
\caption{Next word probability distribution.}
\end{figure}

This goes on until we can compute the probability of the entire sentence. Upon that, the idea of entropy is built.

\textbf{Entropy} (in the context of information theory and not \href{https://en.wikipedia.org/wiki/Entropy}{thermodynamics}) can be explained as the amount of information conveyed by a message. Simply it is the average information carried by each letter in a language \emph{L}. However, since most languages contain infinite amounts of text (unknown distribution), we cannot accurately calculate entropy and thus \emph{cross-entropy} is used to calculate how similar language distributions are.

Moving back to perplexity, it is mathematically defined as $exp(-\frac{1}{t}\sum_{i}^{t}\log p_\theta (x_i|x_<i))$ with $\log p_\theta(x_i|x_<i)$ being the log-likelihood of the $i_\text{th}$ token, given the previous tokens $x_\text{<i}$ and is always equal to $2^\text{Entropy}$. It can be described as the "uncertainty" of the model when predicting the next symbol in the sequence. All models aim to minimize perplexity (even though lower perplexity does not always mean a better model) and the tokenization process directly impacts it's value.

After the first training iteration, a validation iteration occurs with the exact same steps as the training loop with the only difference that our data is now taken from the validation dataloader. This "2-step" process (train loop - validation loop) happens as many times as we have set epochs. Below are graphs for the training and validation losses along with perplexity and average losses values.

In the following plot, we present our training and validation loss across all 3 epochs (we concatenate each epoch into a long list). The plot seems "fuzzy" as there are constant jumps in loss. This is caused because we are using "mini-batches" which are essentially really small batches (remember we set batch-size to 2) that help when training with a GPU. With every batch, there is a chance of "bad" data that cause a loss jump. The important thing here is that notice a clear downward trend as loss slowly converges towards our final \textbf{0.02} value which is expected as this is train loss.

The validation loss also follows the same pattern which is a healthy indicator that our model did not overfit. Below we have split the loss plot into 3 plots, each showing the entirety of our loss values per iteration, every $50_\text{th}$ loss value and every 10th value respectively. This helps to make the downward trend of the loss more clear.

\begin{figure}[H]
    \centering
    \includegraphics[width=\textwidth]{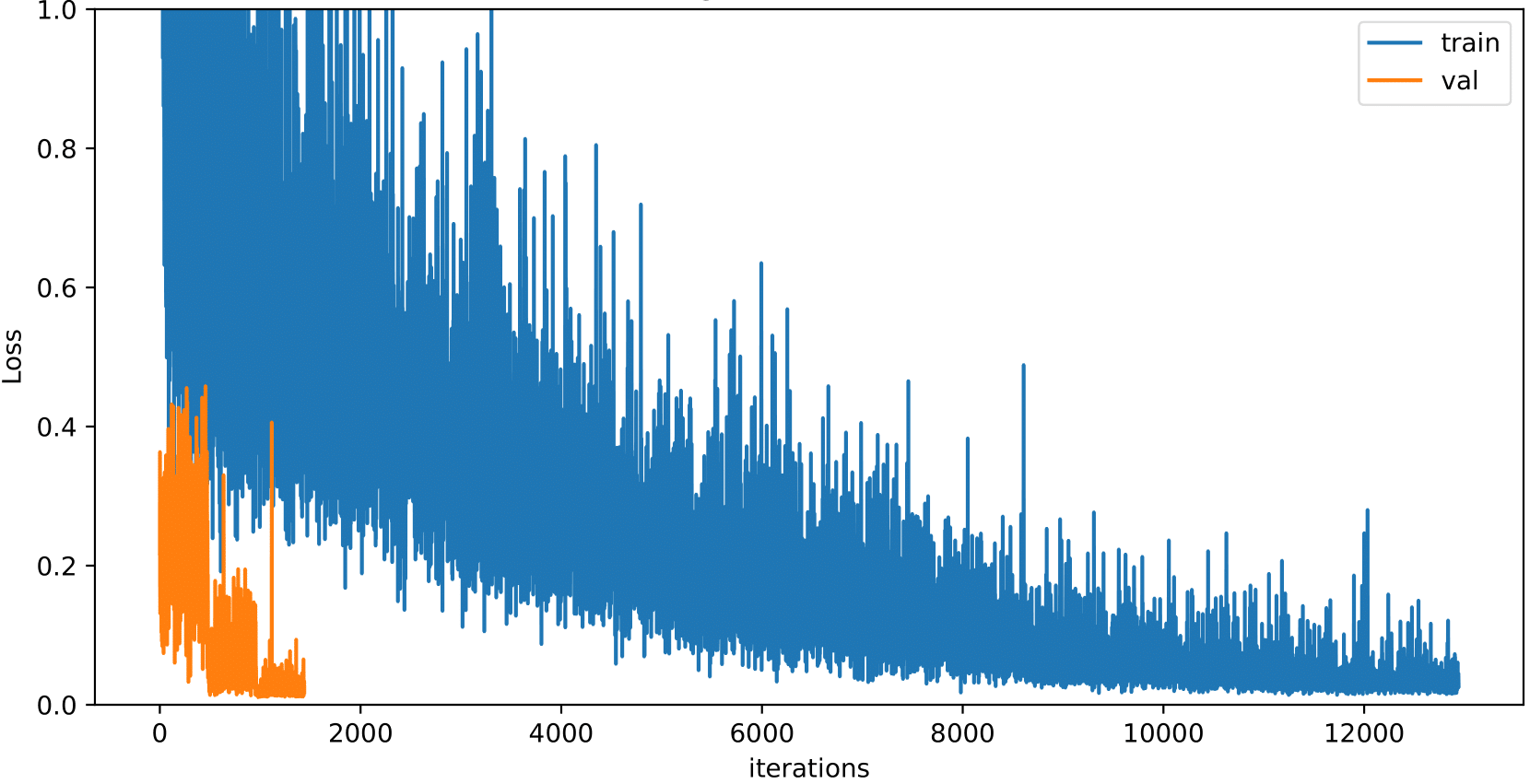}
    \caption{Train/Validation plot}
\end{figure}

\begin{figure}[H]
    \centering
    \includegraphics[width=\textwidth]{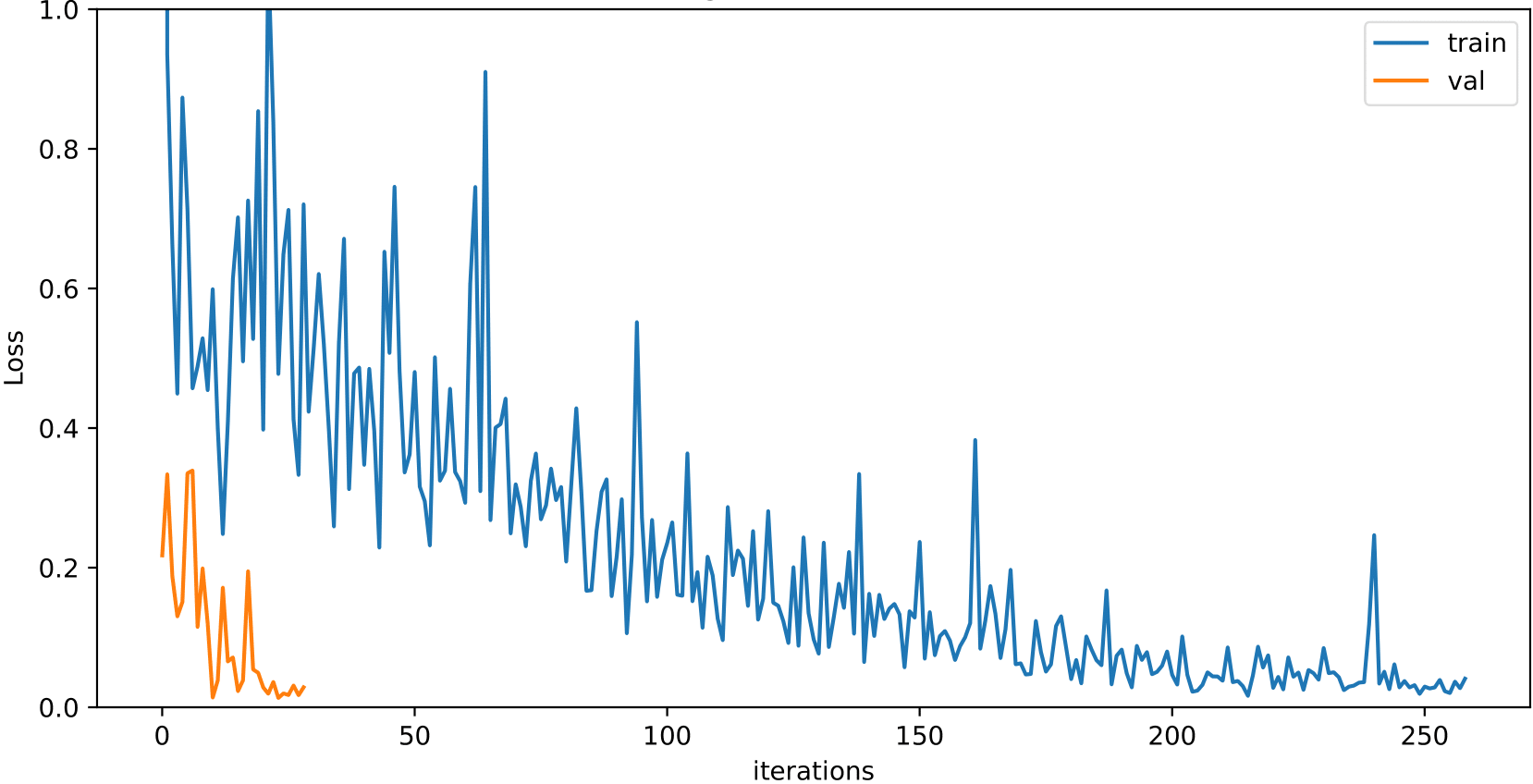}
    \caption{Train/Validation plot (Every $50_\text{th}$ value)}
\end{figure}

\begin{figure}[H]
    \centering
    \includegraphics[width=\textwidth]{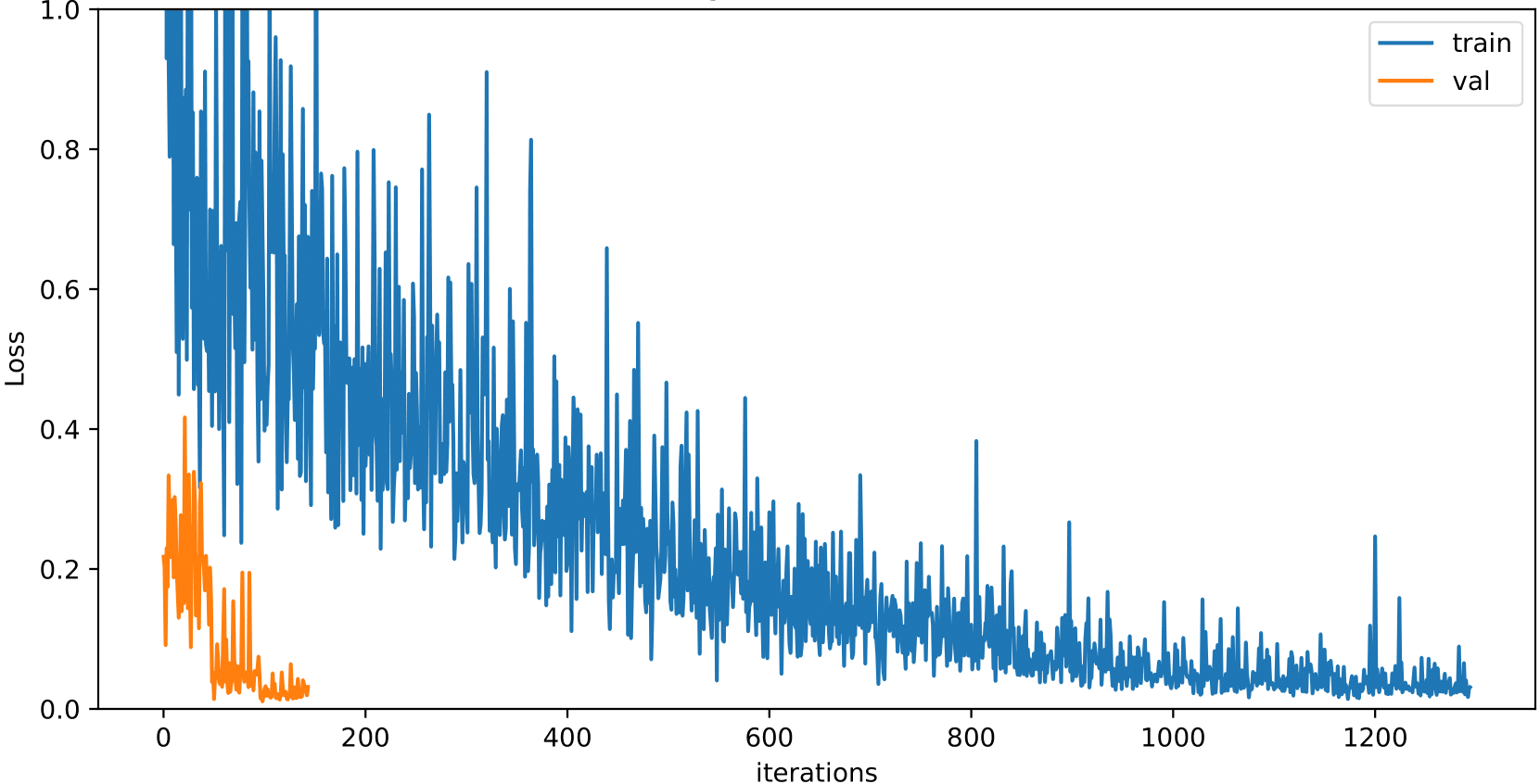}
    \caption{Train/Validation plot (Every $10_\text{th}$ value)}
\end{figure}

\subsection{Testing}
For the Testing phase, a similar approach was followed. First of all we create the test dataloader using the test dataset with the difference that we now use a sequential sampler instead of a random one since we are testing and do not care about shuffling. The test loop consists of the following parts:

\begin{enumerate}
    \item We set the model to evaluation mode using the \emph{model.eval()} method. This informs the model (or rather various layers of the model) that what we are currently doing is evaluation and not training. This means that layers such as \textit{Dropout}, \textit{BatchNorm} etc. will behave differently. Specifically, \textit{BatchNorm} layers will now use running statistics rather than "per-batch" ones and \textit{Dropout} layers will now deactivate.
    \item We get the next batch of data as usually from our \textit{test} dataloader this time (Remember it includes input IDs, labels and attention masks)
    \item At this point, before we do any model output calculation, it is really important to wrap the \emph{model()} call with \textit{torch.no\_grad()}. This sets all the \textit{requires\_grad} flags to False. This flag is set by default to True when tensors are created and basically informs them that every operation that happens involving them, need to be tracked because gradients need to be calculated later in the backward pass (\emph{.backward()} call). Since we will not calculate gradients and are just doing testing we set the flags to False.
    \item Next, as per usual batch losses are calculated along with the total testing loss.
    \item Finally, when the loop ends, the total average loss is calculated along with perplexity.
\end{enumerate}

The final values we get for our model are

Test Loss: \textbf{0.02}\\
Test perplexity: \textbf{1.02}

\subsection{Results}
Once the model finished training, we save our model and now it is ready to produce recipes. By passing "keywords" (ingredients) as the recipe's first word and then the "\textit{<START\_TITLE>}" token, we let the model figure out what the next characters should be and thus create a recipe character by character. Other than the \emph{input\_ids} which are the keywords, we pass the following arguments to the \textit{.generate()} method.

\begin{itemize}
\centering
    \item \textbf{num\_beams} - Number of beams in \textit{Beam search} (We set that to 5).
    \item \textbf{no\_repeat\_ngram\_size} - The size of \textit{n-grams} that will appear only once (We set that to 2).
    \item \textbf{max\_length} - This is the maximum length the generated tokens can have (We set that to 1000).
    \item \textbf{num\_return\_sequences} - How many returned sequences ber element in batch (We set that to 1).
    \item \textbf{eos\_token\_id} - This is the id of the "end of text" token (We pass that through the tokenizer).
\end{itemize}

The first sample recipe we present below has been "prettified" meaning we have removed all tokens that the model created (such as \textit{|<endoftext>|}) and we have formatted it into a readable string.

\clearpage

\begin{center}
\textbf{- Name -}\\
\vspace{0.2cm}
Gluten-Free Vegan Chocolate Pear Cake\\
\vspace{0.2cm}
\textbf{- Ingredients -}\\
\vspace{0.2cm}
\begin{minipage}{0.5\textwidth}
\begin{itemize}
\item 60 ml (or rice milk) almond milk
\item 1 tsp apple cider vinegar
\item 3 ripe peeled cored and chopped pears
\item 10 g vegan melted and cooled dark chocolate
\item 75 g sifted all purpose flour
\item 2 tbsp cocoa powder
\end{itemize}
\end{minipage}\\
\vspace{0.2cm}
\textbf{- Instructions -}\\
\vspace{0.2cm}
\begin{minipage}{0.85\textwidth}
\begin{itemize}
\item[--] Heat the oven to 180°c (160° fan)
\item[--] Grease a 450g loaf tin
\item[--] Melt the milk and vinegar in a pan over a low heat
\item[--] Stir in the apple and set aside
\item[--] Whisk together the flour
\item[--] Cocoa and baking powder and add to the melted chocolate
\item[--] Pour into the tin and smooth the top sprinkle with sift the remaining chocolate and leave to cool
\item[--] Bake for 30-40 minutes until cooked through
\item[--] Cover with foil if the cake is browning too much as it cooks
\item[--] Sprinkle with a little sesame oil and it cool on a wire rack
\item[--] Store wrapped
\item[--] In an airtight tin for at least 2 days before cutting* Tips* If you prefer non-stick cake
\item[--] You could substitute water
\item[--] Yield will depend on type of cake used
\item[--] Can be frozen in air tight containers
\end{itemize}
\end{minipage}
\end{center}

\clearpage
Using the ingredient "\emph{chocolate}" we get the following recipe.

\begin{center}
\textbf{- Name -}\\
\vspace{0.2cm}
Chocolate Whoopie\\
\vspace{0.2cm}
\textbf{- Ingredients -}\\
\vspace{0.2cm}
\begin{minipage}{0.5\textwidth}
\begin{itemize}
\item 110 g butter
\item 175 g sugar
\item 1 each beaten eggs
\item 125 ml milk
\item 3 tbsp cocoa powder
\item 300 g all purpose flour
\item 2 tbsp baking powder
\end{itemize}
\end{minipage}\\
\vspace{0.2cm}
\textbf{- Instructions -}\\
\vspace{0.2cm}
\begin{minipage}{0.85\textwidth}
\begin{itemize}
\item[--] Heat the oven to 170ºc (150º fan) 325ºf
\item[--] Gas mark 5 grease and line two baking sheets
\item[--] Cream together the butter and sugar until light and fluffy then beat in the egg a little at a time mix in 1/3 of the milk and pour into about 16 mounds on the baking sheet
\item[--] Flatten slightly and bake for about 15 minutes or until firm to the touch then transfer to a wire rack and let cool
\item[--] Cut the cakes in half and sprinkle with coarsely ground cocoa
\end{itemize}
\end{minipage}
\end{center}

\clearpage

\begin{center}
\textbf{- Name -}\\
\vspace{0.2cm}
Chocolate banana tartlets with chanterelles\\
\vspace{0.2cm}
\textbf{- Ingredients -}\\
\vspace{0.2cm}
\begin{minipage}{0.5\textwidth}
\begin{itemize}
\item 100 g raisins
\item 55 g chopped candied lemon zest
\item 225 g (candied) quartered cherries
\item 200 g caster sugar
\item 3 beaten eggs
\item 15 ml vanilla extract
\item 250 ml 30\% fat cream
\item 150 g butter
\item 75 g dark chocolate
\item For dusting icing sugar
\end{itemize}
\end{minipage}\\
\vspace{0.2cm}
\textbf{- Instructions -}\\
\vspace{0.2cm}
\begin{minipage}{0.85\textwidth}
\begin{itemize}
\item[--] Place all the fruits in a bowl with the chia seeds
\item[--] Stir well
\item[--] Cover and chill for 2 hours
\item[--] Heat the oven to 160°c (140° fan) gas 3 line a large baking tray with non-stick baking paper
\item[--] Melt the chocolate couverture
\item[--] Beat the egg and sugar until creamy add the milk
\item[--] Vanilla and cream and stir until blended
\item[--] Sieve the mixture into a food processor and pulse until a rough dough comes together
\item[--] Adding more milk 1 teaspoon at a time if the dough seems dry
\item[--] Using clean floured hands
\item[--] Gently fold in the flour and cocoa
\item[--] Drop heaped teaspoons on to the baking trays and bake for 15-20 minutes until golden and puffy
\item[--] Leave to cool for 5 minutes before serving
\item[--] Whisk the cream until thick pipe or spoon on top of the cakes
\item[--] Sprinkle with confectioners' sugar and chocolate shavings and decorate each piece with a white chocolate peel
\end{itemize}
\end{minipage}
\end{center}

\clearpage

\begin{center}
\textbf{- Name -}\\
\vspace{0.2cm}
Frog legs with lemon and thyme \\
\vspace{0.2cm}
\textbf{- Ingredients -}\\
\vspace{0.2cm}
\begin{minipage}{0.5\textwidth}
\begin{itemize}
\item 1 kg / 32 ( or frogs' legs) frogs
\item 2 lemons
\item 3 tbsp olive oil
\item 250 g / 1 (¼-ounce) packet active dry yeast
\item 4 large eggs
\item 5 cups / 200 g grated gruyère cheese
\item 8 slices (for serving) bread
\end{itemize}
\end{minipage}\\
\vspace{0.2cm}
\textbf{- Instructions -}\\
\vspace{0.2cm}
\begin{minipage}{0.85\textwidth}
\begin{itemize}
\item[--] In a pan* Gently melt the lemon
\item[--] Add the oil
\item[--] A little salt and a pinch of sugar
\item[--] Bring to the boil and then cover and cook over a medium heat for 25 minutes
\item[--] Stirring occasionally
\item[--] Until the frogs are soft
\item[--] Remove from the heat
\item[--] Leave to cool slightly and stir in the remaining lemon juice
\item[--] Place the loaf of bread on a work surface sprinkled with flour
\item[--] Sprinkle with a little flour and quickly knead together to make a smooth
\item[--] Even ball of dough
\item[--] Beat the egg whites until stiff
\item[--] Then fold into the mixture with the cheese
\item[--] Roll out the dough to a rectangle measuring 30 x 40 cm
\item[--] Put in a baking dish lined with baking parchment
\item[--] Cover and brush the surface lightly with non-stick baking paper
\item[--] Bake for 10-15 minutes until golden
\end{itemize}
\end{minipage}
\end{center}

At this point, its important to note that most of the recipes generated even though they are similar to existing ones from the dataset, they are not identical. There are ingredient and instruction differences that make them unique. This happens mainly due to the small size of our dataset which did not allow the model to generalize very much. Also, results seem to be worse when using ingredients not found in the dataset. However, no matter what, the model produces proper recipes, with the proper structure and with instructions which include all the ingredients meaning it has learned to understand the structure really well (as was expected since it is a \emph{GPT-2} model) and not so much the content (due to lack of data).

\section{Discussion}
There is a plethora of ideas on how someone could take this entire project one step further. Most of them stem from the creation of additional data as we were "data-bound" and not "model-bound". Meaning that for our project, the absence of usable data (and at times data in general) was what made it extremely hard to properly train our model, and not the lack of good and performant models. That was the main hindrance that stopped us from achieving even better results. Nevertheless, we found out the power of "pre-training" and "fine-tuning" and witnessed the powerful \emph{GPT-2} model in action. Below are a few ways someone could build upon this project and update it, achieving even better results.

First and foremost, with the rise of "fine-dining" as a trend, there could be a bloom of fine-dining websites and places where recipe data could be gathered from. Enriching the dataset we have created will most probably be the catalyst of achieving way better results and creating a solid dataset for future uses. For the time, the only places we found usable and relatable data were the ones we used.

By finding clever and creative ways to augment the already gathered data (Remember the idea of calculating the powerset of ingredients and using that as a base for every recipe) will most probably be the best alternative to the lack of "fine-dining" raw data at least for the present.

Using the new \emph{GPT-3} model or any other transformer might also improve results as we have notice a trend of newer models performing adequately with less and less data as time moves on. Also, using different types of language models in place of autoregressive ones like the one we have used might prove fruitful. The ideas of model types like \emph{Autoencoding}, \emph{Multi-modal}, \emph{Seq-to-Seq} and more could be a better alternative than what we used here.

Nevertheless, this entire project serves only as a precursor to what is possible given today's advancements in deep learning and it could serve as a solid base for something bigger in the (near) future.

\section{Conclusion}
In conclusion, while our results might have been somewhat inconsistent and the training process a bit rocky, we saw that with relatively minimal data \emph{GPT-2} managed to achieve clear and acceptable results. The model clearly understood the structure of the recipes as well as the relations between each category (Title, instructions and ingredients) and generate novel recipes from a really small dataset. Our final testing loss was \textbf{0.02} and our final perplexity was \textbf{1.02} with a train time of $\approx$85 mins on a \emph{NVIDIA Tesla P100 GPU}.

\bibliography{sample}

\end{document}